%% file: paper.tex
\documentclass[11pt]{article}
\usepackage[preprint]{acl}
\usepackage{times}
\usepackage{latexsym}
\usepackage[T1]{fontenc}
\usepackage[utf8]{inputenc}
\usepackage{microtype}
\usepackage{inconsolata}
\usepackage{graphicx}
\usepackage{booktabs}
\usepackage{amsmath,amssymb}
\usepackage{tabularx}
\usepackage{array}
\usepackage{placeins}
\usepackage{url}
\usepackage{capt-of}

\title{RCCM: Diagnosing Gains and Harms in Reasoning-Prompt Transfer}
\author{Yuqi Zhang \quad Di Zhang}
\hypersetup{
  pdftitle={RCCM: Diagnosing Gains and Harms in Reasoning-Prompt Transfer},
  pdfauthor={Yuqi Zhang and Di Zhang}
}
\begin{document}
\maketitle
\raggedbottom
\begin{abstract}
A reasoning prompt can improve one deployment and degrade another. RCCM (Reasoning-Control Capability Mismatch) examines these reversals by pairing answers to the same questions with records of how generation ends. Across five open-weight deployments and nine strategies, aggregate accuracy hides substantial question-level gains and harms; a common-budget MATH500 evaluation examines transfer on a separate sample. On 60 additional questions, concise reasoning improves Phi's automatic answer matching at a 2,048-token cap across three collections, but sharply reduces Qwen's. Author-reviewed final-answer commitments give a smaller, statistically unresolved Phi gain. Reserving an answer stage also improves matching, with additional input and call costs. These results show why a prompt benefit must be interpreted alongside its execution and evaluation: matching a reference answer need not mean delivering it, and an apparent compression benefit may admit another explanation. RCCM records this evidence for reproducible transfer comparisons.
\end{abstract}

\section{Introduction}
Transferring a successful reasoning prompt to another deployment does not necessarily preserve its benefit. An advantage can reverse across models \citep{wang2025promptbridgecrossmodelprompttransfer}, and prompt choice and formatting affect performance even within a model \citep{mizrahi-etal-2024-state,ICLR2024_6c0e99d7,zhuo-etal-2024-prosa}. With reasoning prompts, changing the text also changes the requested procedure: shorten a derivation, organize it into stages, or check an earlier answer. To understand the resulting accuracy change, we need to know which answers improve and which become wrong.

Consider Chain of Draft (CoD), which asks for terse intermediate steps \citep{xu2025chaindraftthinkingfaster}. On the same 120 development questions, our Phi deployment answers 83 correctly with Chain of Thought (CoT) and 98 with CoD. Qwen moves in the opposite direction, from 88 to 56. The paired answers reveal how those totals arise: CoD rescues 20 CoT errors and harms five correct answers on Phi, but rescues five and harms 37 on Qwen (Figure~\ref{fig:paired}). The same strategy therefore plays opposite roles. Its net effect depends on the gains and harms it produces under each execution configuration.

These strategies impose different demands on generation. CoT requests intermediate steps \citep{NEURIPS2022_9d560961}; CoD requests brief drafts \citep{xu2025chaindraftthinkingfaster}; checking and repair request inspection and revision \citep{NEURIPS2023_91edff07,ICLR2024_8b4add8b}. A requested operation, however, is not its observed execution, and execution is not task success. A shorter response may still be wrong; a correctly extracted number may still be an intermediate candidate. Understanding transfer requires following the requested operation through the response to the answer being evaluated.

RCCM studies \emph{reasoning-control capability mismatch} through the requests a strategy makes and the outcomes a deployment produces. A deployment comprises the checkpoint, chat wrapper, decoding policy, output budget, and answer parser. For each fixed strategy comparison, we pair responses to common questions and relate answer changes to execution events. The resulting record shows which questions change correctness, how generation ends, and what the evaluator accepts as an answer.

The Phi observations motivate two tests of the apparent compression benefit: a concise request on additional questions and an explicit answer stage under the same generated-token cap. We also review whether the original responses commit to a correct final answer. These comparisons distinguish automatic matching from answer delivery and test explanations beyond brevity.

We make three contributions. First, a paired execution model decomposes a prompt's net benefit into execution frequencies and the gains and harms within each category. Second, experiments across deployments expose strategy reversals and test a compression benefit using repeated collections, reviewed answer commitments, and answer-delivery controls. Third, a reproducible transfer record specifies the strategy, execution configuration, evaluator, costs, and validation scope, so the comparison can be checked and reused.

\section{Related Work}
\paragraph{Designing and adapting reasoning.}
CoT, self-consistency, and self-refinement alter reasoning or answer aggregation \citep{NEURIPS2022_9d560961,wang2023selfconsistency,NEURIPS2023_91edff07}. CoD and Sketch-of-Thought use compact representations \citep{xu2025chaindraftthinkingfaster,aytes-etal-2025-sketch}; decoding can also expose reasoning paths without explicit CoT instructions \citep{NEURIPS2024_7a8e7fd2}. Format-Adapter generates and selects reasoning formats under a test-time budget \citep{wang-etal-2026-format}. Here, the procedure is fixed: we study how its answers and execution change when it is used on another deployment.

\paragraph{Compression and budget control.}
Concise CoT reports model-dependent accuracy costs \citep{renze2024concise}, while reasoning-step experiments vary demonstration granularity \citep{jin-etal-2024-impact}. \citet{lee2025tokencomplexity} examine 31 compression prompts, question-level length thresholds, and accuracy--length tradeoffs. TALE estimates per-question token budgets \citep{han-etal-2025-token}; Mid-Think combines token-level cues to induce intermediate reasoning budgets \citep{yang-etal-2026-mid}. Strict-cap evaluation compares truncation with an early-stop answer phase \citep{sun-etal-2025-empirical}, and budget-aware strategy evaluation accounts for inference costs \citep{wang-etal-2024-reasoning-token}. Our comparison follows a fixed prompt edit on common questions, linking each rescue or harm to paired stopping states. An additional question sample and two hard caps test whether the development pattern persists.

CLARO combines attribute-guided prompting with reinforcement learning \citep{schlesinger-etal-2026-claro}; Budget Guidance trains a remaining-length predictor to steer generation \citep{li-etal-2026-steering}. \citet{zhou-etal-2026-thinking} analyze positive and negative answer flips as reasoning budgets increase. REFRAIN adjusts stopping thresholds online using reflective-redundancy signals, without fine-tuning the reasoning model \citep{sun-etal-2026-stop}. RCCM relates answer flips to execution changes under a fixed prompt edit. This joint analysis, followed by evaluator and delivery controls, distinguishes the study from budget-controller design.

\paragraph{Sensitivity and transfer.}
Multi-prompt and sensitivity studies document dependence on prompt choice \citep{mizrahi-etal-2024-state,ICLR2024_6c0e99d7,lu-etal-2024-prompts,zhuo-etal-2024-prosa}; matching rules also affect measured sensitivity \citep{hua-etal-2025-flaw}. PromptBridge learns a mapping from calibrated source--target prompt pairs \citep{wang2025promptbridgecrossmodelprompttransfer}, while Beyond the Answer compares scientific prompting across models \citep{pandey2026answerdecodingbehaviorllms}. Instance-adaptive CoT addresses instance-level harm \citep{NEURIPS2024_e304e04a}. Our comparison keeps the prompt edit fixed and tests how its benefit changes with execution conditions, without learning a cross-model mapping.

\paragraph{Control requirements and task outcomes.}
The Format Tax separates prompt-level format costs from decoder constraints, analyzes answer flips and length, and tests decoupled reasoning and formatting \citep{lee2026formattax}. We share its distinction between an execution requirement and answer correctness. RCCM measures the joint stop-state and answer-transition distribution for a fixed edit and hard cap, then checks whether answer matching, post-thought scope, and reviewed answer commitment support the same interpretation. We distinguish reference-assisted repair from intrinsic feedback \citep{ICLR2024_8b4add8b} and observable responses from the computation that produces them \citep{NEURIPS2023_ed3fea90}.

Procedural-execution tests assess adherence to prescribed arithmetic steps \citep{panda2026procedural}. Our responses are freely generated, so a stop state cannot establish that the requested intermediate steps were followed. Human-aligned multiple-choice evaluation likewise distinguishes automatic matches from answer judgments \citep{molfese-etal-2025-right}. For open answers, we keep matching, structural scope, and commitment review as separate endpoints.

\section{From Control Requests to Answer Changes}
\label{sec:framework}
\paragraph{A paired model of execution.}
Fix baseline and target strategies $m_0,m_1$, including their templates and any multi-call procedure. Draw a question $X$ from a declared distribution $Q$, which may be uniform over a fixed benchmark. Under deployment $e$, let $(O_0,O_1)$ be the paired response records, including stop reasons and interface failures. Their conditional joint law $K_e(\cdot\mid X)$ describes the collection protocol, with deterministic generation as a special case. A fixed evaluator $s(X,O_j)\in\{0,1\}$ assigns the answer-match label $Y_j$. An interface failure yields no successful match; the latent answer's mathematical correctness remains unknown. RCCM models these observable records rather than an unmeasured internal capacity.

\paragraph{Gains, harms, and execution.}
Write $D=Y_1-Y_0$. Its expectation $\delta_e$ is the target's advantage, in probability units. The events $D=1$ and $D=-1$ are \emph{rescues} and \emph{harms}. On $n$ common questions their counts give
\begin{equation}
\widehat\delta_e=(R-H)/n,\qquad
\Delta\mathrm{Accuracy}=100\widehat\delta_e\ \text{(pp)}.
\label{eq:paired}
\end{equation}
They compare responses to the same question, not necessarily a revision of the same response. Let $Z$ be an exhaustive execution category shared across deployments; our analysis uses the pair of stop states (length, normal, or interface failure). Define $w_e(z)=\Pr_e(Z=z)$ and $\mu_e(z)=\mathbb{E}_e[D\mid Z=z]$. Then
\begin{equation}
\delta_e=\sum_z w_e(z)\mu_e(z).
\label{eq:joint}
\end{equation}
The contribution of an execution category depends on how often it occurs and whether its answer changes are beneficial. Table~\ref{tab:joint} gives the empirical decomposition.

\paragraph{What must persist for a benefit to transfer?}
Consider another deployment $e'$ with the same strategies, question distribution, evaluator, and category definitions. Put $v_z=\min\{w_e(z),w_{e'}(z)\}$ and $\mathrm{TV}(w_e,w_{e'})=\frac12\sum_z|w_e(z)-w_{e'}(z)|$. An elementary bound is
\begin{equation}
\begin{split}
|\delta_{e'}-\delta_e|\ \le\;&2\mathrm{TV}(w_e,w_{e'})\\
&+\sum_z v_z|\mu_{e'}(z)-\mu_e(z)|.
\end{split}
\label{eq:transfer}
\end{equation}
Terms with $v_z=0$ contribute zero. The bound separates shifts in execution frequency from shifts in conditional task benefit. Identical stopping distributions alone cannot preserve a benefit: every pair could stop normally while changing from all rescues to all harms. Both deployments' joint outcomes are needed to evaluate the bound; its purpose is to account for an observed transfer difference, not to predict one from source records alone.

Appendix~\ref{app:theory} proves the bound and covers absent categories, evaluator changes, and sampling assumptions. We estimate execution frequencies and conditional gains and harms, then examine evaluator sensitivity. Stopping is observed after generation; the bound does not identify its causal effect. The additional-question study tests the fixed edits motivated by these measurements.

\section{Experimental Setup}
\label{sec:setup}
\input{blocks_table}
\paragraph{Deployments and data.}
The development panel runs Qwen 2.5 7B Instruct, Gemma 3 4B, Llama 3.2 3B Instruct, Mistral 7B, and Phi-4 Mini Reasoning 3.8B through Ollama. Its 120 questions comprise 40 GSM8K problems \citep{cobbe2021trainingverifierssolvemath} and two 40-question MATH strata chosen using pilot difficulty \citep{hendrycks2021math}. We set temperature to zero and supply no explicit system message; deployment defaults may still apply. Output caps are 2,048 tokens for Qwen/Gemma, 1,024 for Llama/Mistral, and 4,096 for Phi. Appendix~\ref{app:templates} separates retained historical identities from newer runtime snapshots and identifies missing serving metadata.

MATH500 evaluates six candidates on Qwen, Gemma, and Llama at a common 2,048-token cap. None of its 500 question texts exactly matches the development panel; sources and indices follow MATH/PRM800K \citep{hendrycks2021math,ICLR2024_aca97732}. Diagnostic probes use 60 selected development questions, 20 per stratum. A subsequent study uses a separate fixed 60-question sample after excluding exact matches to recorded local inputs. Table~\ref{tab:blocks} distinguishes these question sets from repeated generations and configurations.

\paragraph{Strategies.}
The nine development strategies are CoT, Concise CoT, CoD, Key-Step Draft (KSD), Full-Retry, KSPAC-Repair, Contrastive Key-Step Prompting (CKP), Meta-Cognitive Router (MCR-Route), and Progressive Key-Step Reasoning (PKR). CoT and CoD instantiate the published ideas under our settings. The other names identify local templates or workflows. MCR-Route is a development-calibrated routing reference. MATH500 uses CoT/Concise/CoD/KSD and two single-generation variants, KSPAC-Check and MCR-Instruction, rather than the development repair and routing pipelines. Appendix~\ref{app:templates} gives the exact bindings.

\paragraph{Scoring and inference.}
The primary score is answer matching after the archived extraction procedure. We retain raw text, extraction provenance, and stop reasons. Exact paired McNemar tests \citep{mcnemar1947,fagerland2013} receive Holm correction \citep{holm1979} over all 36 pairs per nine-strategy deployment and all 15 pairs per MATH500 deployment. A \emph{statistical top band} contains candidates not significantly separated from the point winner in that family, without asserting equivalence. Pointwise 95\% paired bootstrap intervals \citep{efron1979} use 20,000 resamples of common questions. Numerical \emph{near-best} is a separate criterion: within $\max(2,\lceil0.03n\rceil)$ matches of the observed best. Original collections contain one generation per request; later repeated collections retain the same questions and do not enlarge the independent sample.

\section{How Fixed Strategies Change Roles}
\begin{figure*}[t]
\centering
\includegraphics[width=\textwidth]{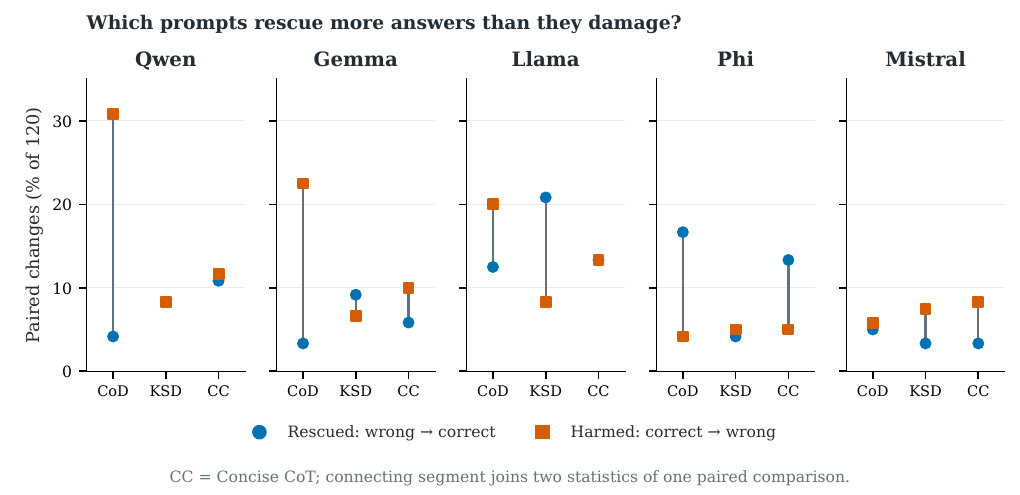}
\caption{Paired gains and harms relative to CoT on 120 development questions per deployment. CC denotes Concise CoT. Blue circles show baseline errors rescued by the target; orange squares show correct baseline answers harmed. Their directed difference is the net accuracy change, not a confidence interval. Deployment-specific caps are given in Section~\ref{sec:setup}.}
\label{fig:paired}
\end{figure*}

\paragraph{Opposite effects under the same request.}
Table~\ref{tab:dev} reports all nine strategies. CoD improves Phi by 15 correct answers and reduces Qwen by 32. Llama benefits from KSD, moving from 69 to 84 correct answers, while Phi moves from 83 to 82. These paired effects within each deployment cannot be summarized by the models' overall accuracy ordering.

\begin{table}[t]
\centering\small
\setlength{\tabcolsep}{3.5pt}
\begin{tabular}{lrrrrr}
\toprule
Strategy & Qwen & Gemma & Llama & Mistral & Phi\\
\midrule
CoT & \textbf{73.3}&70.0&57.5&10.0&69.2\\
Concise &72.5&65.8&57.5&5.0&77.5\\
CoD &46.7&50.8&50.0&9.2&\textbf{81.7}\\
KSD &\textbf{73.3}&72.5&\textbf{70.0}&5.8&68.3\\
Full-Retry &60.0&66.7&55.0&\textbf{19.2}&60.0\\
KSPAC-Repair &60.0&66.7&50.8&16.7&64.2\\
CKP &72.5&67.5&60.8&15.8&77.5\\
MCR-Route &69.2&\textbf{75.0}&68.3&13.3&72.5\\
PKR &67.5&65.8&37.5&6.7&63.3\\
\bottomrule
\end{tabular}
\caption{Development answer accuracy (\%), $n=120$ per cell. Bold marks a point maximum. MCR-Route uses development information; its score is not unseen-data routing performance.}
\label{tab:dev}
\end{table}

The rankings in Table~\ref{tab:dev} also differ. Mean pairwise Spearman correlation among Qwen, Gemma, Llama, and Phi is 0.394, ranging from $-0.139$ for Gemma--Phi to 0.831 for Gemma--Llama. Mistral's scores are 5.0--19.2\%, providing a low-performing reference; those scores alone do not diagnose an interface fault.

\paragraph{Equal accuracy hides different answers.}
On Qwen, CoT and KSD both score 88/120, but KSD rescues ten errors and harms ten correct answers. Twenty questions change correctness without changing accuracy. On Llama, CoD rescues 15 CoT errors but introduces 24 harms, giving a net loss. Figure~\ref{fig:paired} shows these cancellations. Such complementary answers offer an opportunity for selection, provided a selector can recognize them without access to the reference.

\paragraph{A common budget on MATH500.}
At the common 2,048-token cap (Table~\ref{tab:math}), CoD trails CoT on all three deployments. Qwen and Llama score highest with CoT, Gemma with KSPAC-Check. Gemma's 4.0-point Check advantage over CoT is not separated after Holm correction ($p=0.1059$); both remain in its statistical top band. Qwen's band excludes only CoD; Llama's excludes CoD and Concise CoT.

\begin{table}[t]
\centering\small
\begin{tabular}{lrrr}
\toprule
Strategy & Qwen & Gemma & Llama\\
\midrule
CoT &\textbf{72.6}&70.8&\textbf{42.0}\\
Concise CoT &69.8&68.6&34.6\\
CoD &39.4&50.6&24.6\\
KSD &67.8&68.0&37.8\\
KSPAC-Check &71.4&\textbf{74.8}&41.4\\
MCR-Instruction &69.4&69.4&40.2\\
\bottomrule
\end{tabular}
\caption{MATH500 accuracy (\%), $n=500$ per cell, common 2,048-token cap. Check and Instruction are single-generation variants, distinct from the development pipelines.}
\label{tab:math}
\end{table}

The six-candidate oracle scores 86.0\%, 84.8\%, and 68.2\% on Qwen, Gemma, and Llama, exceeding the best fixed scores by 13.4, 10.0, and 26.2 points. It measures complementarity: selection cannot use the reference answer.

\paragraph{Matching and sampling checks.}
Replaying extraction and matching reproduces the archived correctness labels for all 5,400 development and 9,000 MATH500 responses. Two extracted strings differ, but their correctness labels do not. An additional normalization check converts seven development errors to matches; the point winners are unchanged, apart from a new Qwen tie. Appendix~\ref{app:sampling} reports a separate five-sample plurality analysis with its ties and generation settings. These checks examine how extraction, normalization, and decoding choices affect the comparison.

\section{Execution, Evaluation, and Answer Delivery}
\label{sec:execution}
\subsection{What changes in the diagnostic probes?}
On Phi's 60 selected development probes at a 2,048-token cap, the concise request scores 40/60 versus 28/60 for full CoT. It rescues 14 errors and harms two correct answers: $+20.0$ points, 95\% interval $[8.3,31.7]$, Holm $p=0.0293$ across the seven reported probe comparisons. Median emitted tokens fall from 2,048 to 1,106; actual length stops fall from 34 to 21.

Table~\ref{tab:joint} shows where this gain occurs. Thirteen questions move from a length stop to a normal stop, and 12 also move from wrong to correct. The 21 length-to-length pairs contain two rescues and two harms, canceling in the net score. Equation~\ref{eq:joint} recovers the overall gain as $(13/60)(12/13)=0.20$. Its concentration in length-to-normal pairs motivates testing a shorter request on additional questions. Answer-only prompting gives 33/60 with a median 1,445.5 tokens, further separating the requested brevity from the observed response length.

\begin{table}[t]
\centering\small
\begin{tabular}{lrrrrr}
\toprule
Full $\to$ concise & $n$ & $R$ & $H$ & Both & Neither\\
\midrule
Length $\to$ normal &13&12&0&0&1\\
Length $\to$ length &21&2&2&1&16\\
Normal $\to$ normal &26&0&0&25&1\\
Normal $\to$ length &0&0&0&0&0\\
\midrule
Total &60&14&2&26&18\\
\bottomrule
\end{tabular}
\caption{Phi development probes: stop-state and answer transitions. Both/Neither refer to correctness in both/neither condition. Row categories are exhaustive; stopping and correctness are different measurements.}
\label{tab:joint}
\end{table}

Structure and checking probes show numerical gains from multi-stage requests on Qwen and self-check on Gemma, but Gemma declines under repair. Appendix~\ref{app:probes} lists all seven comparisons; only Phi full-to-concise passes their common correction. The nine-condition profiles are marginal diagnostics, not isolated causal effects.

\subsection{Additional questions and two budgets}
\label{sec:confirmation}
We next fixed 60 additional questions, two deployments (Qwen and Phi), four prompts, and two budgets, yielding 960 assignments. The sample comprises 20 GSM8K, 20 MATH levels 1--3, and 20 MATH levels 4--5, selected by seeded within-stratum shuffling and recorded exact-text exclusions (Appendix~\ref{app:selection}). Local registration specifies Phi full-to-concise at 2,048 tokens as the primary comparison, with all eight pair/deployment/budget contrasts in one Holm family. Exact requests, input hashes, model digests, and runtime settings were recorded before inference.

The study produced 957 responses and three terminal interface failures. A documented collection amendment preserved existing responses and completed missing assignments without changing the templates, questions, or scorer. We report answer matches per assigned question, list interface failures separately, and provide missing-answer bounds in the artifact. All three failures occur in Phi's main CoD arm, outside the primary pair. The main and probe pairs retain different historical newline encodings; we therefore analyze them as specified prompt pairs, not a two-factor design.

\paragraph{Matching across collections.}
Phi improves from 37 to 45 matches at 2,048 tokens ($+13.3$ pp; Table~\ref{tab:confirmation}). Seven rescues coincide with length-to-normal transitions; the eighth reverses that stop direction. At 4,096 tokens, the first collection has three rescues and three harms. Its descriptive budget contrast is $-13.3$ pp, interval $[-23.3,-3.3]$, outside the corrected primary tests. Two further same-setting full/concise collections add 960 responses on these questions. Phi's 2,048-token gains are $+18.3/+23.3$ pp (new eight-way Holm $p=.0382/.00049$); at 4,096 they are $+11.7/+8.3$ pp and do not pass correction. Qwen loses matches in every collection (Appendix~\ref{app:repeatability}). The larger cap thus changes the observed effect without consistently eliminating it.

\paragraph{Requested compression and observed execution.}
The concise prompt asks for at most three short reasoning lines. Across the two budgets, Qwen emits a median 4--4.5 tokens, compared with about 936--941 for Phi. Qwen meets a whole-response three-line count on 57/60 questions at each budget; Phi, whose returned content includes marked thought text, meets it on none. The same instruction produces markedly different responses. Phi's improved matching is therefore not evidence of literal compliance with the three-line request.

\paragraph{Three endpoints on the same answers.}
\input{endpoints_table}
Table~\ref{tab:endpoints} applies three criteria to the same responses. The frozen scorer can match a number inside marked thought text. Structural scope considers unmarked output or the suffix of one closed thought section; missing or ambiguous scope counts as no observable match. For answer commitment, one author verified AI-assisted annotations with method names hidden. The resulting gain is smaller and statistically unresolved. All binary completions of the four unknown labels give $[+3.3,+10.0]$ pp; this is a label-uncertainty range, not a confidence interval. The artifact supplies the rubric and blinded inputs. This review does not systematically assess the derivations (Appendix~\ref{app:author_annotation}).

\paragraph{Repeated-measure uncertainty.}
\input{cluster_table}
Across the three chat collections, Phi's mean matching change is $+18.3$ pp at 2,048 and $+6.7$ at 4,096; Qwen's is $-50.6/-56.7$ pp. Table~\ref{tab:cluster} keeps each question's repeated observations together. It separates question-cluster uncertainty from a descriptive sensitivity that also resamples the three collections. These retrospective summaries exclude the raw answer-phase block and leave the registered tests and independent question count unchanged (Appendix~\ref{app:consolidation}).

\subsection{Compression versus answer delivery}
To test whether the benefit is specific to a shorter request, we compare it with reserving room for an answer. Following \citet{sun-etal-2025-empirical}, two answer-phase controls reserve 128 tokens within the same generated-token cap, with or without a warning. A length stop triggers continuation; the controller can close an unmatched thought marker and supply an answer slot. All four strategies run through the same native raw interface on the original 60 questions. Generated-token caps are matched, but total computation is not.

\input{delivery_table}

At 2,048 tokens, both controls improve Phi's matching over full after correction across all 24 comparisons (Table~\ref{tab:delivery}). Their differences from concise do not pass correction. Neither full-to-control contrast passes at 4,096, and neither control shows a corrected gain over full on Qwen. The three automatic scoring endpoints agree on this block; these new outputs have not received the separate commitment review (Appendix~\ref{app:answer_phase}).

Each Phi/2,048 control rescues eleven questions on a two-call target path and two on a one-call path. A has one two-call harm; W has none. These descriptive partitions do not isolate continuation: the arms also change first-stage allowance, injected text and possibly the warning. All paths and budgets are reported in Appendix~\ref{app:consolidation}. Historical fitting and prompt revisions remain auxiliary evidence (Appendix~\ref{app:historical_fit}).

\section{An RCCM Reporting Record}
\label{sec:guard}
An RCCM record documents the comparison behind a transfer claim: the strategy, deployment, evaluator, costs, and validation scope, together with the observed execution and answer changes. It gives readers the evidence for a recommendation without prescribing how candidates must be selected.

\paragraph{Propose an edit from the joint evidence.}
Phi's development gains concentrate in length-to-normal transitions (Table~\ref{tab:joint}), suggesting a concise request under a tighter cap. The edit, comparator, questions, budgets, and evaluator are then fixed. The probe motivates the test; it does not settle the recommendation.

\paragraph{Test the interpretation before adoption.}
Phi's matching gains recur under the tested settings, but commitment review gives a smaller estimate, and an answer stage also improves matching. Recording these comparisons with their gains, harms, uncertainty, and costs lets a reader distinguish evidence that an edit helps from evidence for why it helps.

\paragraph{A worked record.}
In a fixed Qwen split, Concise CoT scores 27/30 on the probe and CoT 24/30. On the remaining 470 questions, CoT scores 339 and Concise 322. The 17-answer gap exceeds the declared 15-answer near-best tolerance. The record retains both the local win and the different validation outcome, rather than treating the probe winner as already transferable.

The historical RCCM-Guard candidate audit illustrates why these tasks should remain separate (Appendix~\ref{app:guard}). Matched-size Direct has a higher near-best hit rate, while Guard more often retains the highest point estimate. These results describe a trade-off, not established selection or validation-cost superiority. The reporting record can accompany either rule.

\section{Conclusion}
RCCM examines prompt transfer through paired answer changes and execution records. The same strategy can rescue some answers and damage others, even when aggregate accuracy is unchanged. Phi's concise request illustrates the further distinction between a matching gain and its explanation: repeated collections, commitment review, and an answer-stage control each change how the benefit should be interpreted. Reporting these results together makes clear which procedure was tested and what evidence supports its reuse.

\FloatBarrier
\section*{Limitations}
The study covers English mathematics and the specified open-weight deployments and templates. Repeated collections and raw controls reuse the same 60 questions. Commitment annotation covers the original Phi pair through author-verified single review, not the later outputs or all rationales. Retained historical identities and current runtime snapshots describe their respective collections. The findings apply to the reported evaluators and execution configurations.

\section*{Ethical Considerations}
We use public mathematical benchmarks and model-generated answers, without recruiting external participants or collecting personal information. Sources and licenses are retained in the artifact. AI tools assisted research, coding, drafting, translation, and initial answer annotation; one author reports verifying every annotation individually. This is not an independent double review or rationale certification. The authors remain responsible for the scientific claims, references, and submission. The reporting workflow does not certify high-stakes use.

\bibliography{references}
\FloatBarrier
\appendix
\input{appendix}
\input{theory_appendix}
\input{followup_tables}
\input{repeat_annotation_tables}
\input{consolidation_appendix}
\input{answer_phase_tables}
\end{document}

%% file: blocks_table.tex
\begin{table*}[t]\centering\small
\begin{tabular}{lrrrr}
\toprule
Block & Interface & Items & Models & Outcomes\\\midrule
Development & Historical & 120 & 5 & 5,400\\
MATH500 & Historical & 500 & 3 & 9,000\\
Additional study & Chat & 60 & 2 & 960 assigned\\
Repeated collections & Chat & Same 60 & 2 & 960 served\\
Answer-phase controls & Native raw & Same 60 & 2 & 960 logical / 1,049 calls\\
\bottomrule\end{tabular}
\caption{Main evidence blocks. The last three share one 60-question set, not 180 independent questions. The original additional study has 957 served outputs and three terminal interface errors. Raw controls are analyzed separately from chat collections. Historical configurations are not replaced by current model identities.}\label{tab:blocks}
\end{table*}

%% file: endpoints_table.tex
\begin{table*}[t]\centering\small
\begin{tabular}{lrrrrr}
\toprule
Endpoint & Full & Concise & R/H & Change (pp) & Inference\\\midrule
Frozen matching & 37 & 45 & 8/0 & +13.3 & 8-way Holm $p=.03125$\\
Structural answer scope & 32 & 44 & 12/0 & +20.0 & Sensitivity Holm $p=.00293$\\
Confirmed equivalent commitment & 44 & 48 & 6/2 & +6.7 & Exploratory $p=.2891$\\
\bottomrule\end{tabular}
\caption{Three endpoints on the original Phi/2,048 full-to-concise pair, $n=60$. Scope uses thought boundaries; commitment uses author-verified AI-assisted labels, with four uncertainties retained. Each endpoint has its own inference; none replaces the registered test.}\label{tab:endpoints}
\end{table*}

%% file: cluster_table.tex
\begin{table*}[t]\centering\small
\begin{tabular}{lrrr}
\toprule
Deployment / cap & Run-average (pp) & Question-cluster interval & Question + collection interval\\\midrule
Phi / 2048 & $+18.3$ & $[11.1,26.1]$ & $[8.9,28.9]$\\
Phi / 4096 & $+6.7$ & $[0.0,13.3]$ & $[-2.2,16.7]$\\
Qwen / 2048 & $-50.6$ & $[-63.3,-37.8]$ & $[-63.3,-37.2]$\\
Qwen / 4096 & $-56.7$ & $[-68.3,-44.4]$ & $[-68.9,-43.9]$\\
\bottomrule\end{tabular}
\caption{Retrospective summaries of three chat collections on the same 60 questions. Pointwise 95\% stratified bootstrap intervals keep paired observations together. The first conditions on the observed collections; the second also resamples their indices, a sensitivity based on only three collections. Raw controls are excluded; registered tests remain unchanged.}\label{tab:cluster}
\end{table*}

%% file: delivery_table.tex
\begin{table*}[t]\centering\small
\begin{tabular}{lrrrrrr}
\toprule
Arm & Matches & R/H vs F & Holm $p$ & Calls & Output tok. & Input tok.\\\midrule
F: full & 38 & -- & -- & 60 & 88991 & 6393\\
C: concise & 41 & 6/3 & $1$ & 60 & 70200 & 6513\\
A: full then answer & 50 & 13/1 & $0.031128$ & 86 & 88588 & 59556\\
W: warned then answer & 51 & 13/0 & $0.0043945$ & 89 & 86728 & 67369\\
\bottomrule\end{tabular}
\caption{Phi/2,048 answer-phase comparison, $n=60$ per arm. The native raw interface and generated-token cap are shared, not total compute. A/W reserve 128 tokens and continue only after a length stop. R/H and Holm $p$ compare each arm with F in the full 24-test family. Commitment review does not cover these outputs.}\label{tab:delivery}
\end{table*}

%% file: appendix.tex
\section{Data Blocks, Templates, and Deployment Records}
\label{app:templates}
This preprint includes a size-limited reproducibility subset in the arXiv ancillary directory: replay code, historical summary tables, and complete confirmation, repeatability, and answer-phase ledgers. The accompanying \texttt{ARXIV\_SUBSET\_README.md} lists included files and runnable checks. Long historical response and prompt ledgers remain in the separate full artifact; they are not included in this arXiv submission. Original records and frozen hashes in the supplied subset are unchanged.

\begin{table*}[t]
\centering\small
\begin{tabular}{lp{6.4cm}p{5.2cm}}
\toprule
Block & Design & Role\\\midrule
Development &120 questions, five deployments, nine strategies & Paired gains/harms and rankings\\
MATH500 &500 questions, three deployments, six candidates; 2,048-token cap & Fixed-candidate evaluation on another set\\
Nine-condition profile &60 questions, four deployments, nine conditions & Marginal behavior under requests\\
Targeted probes &60 selected development questions per tested deployment & Joint stop/answer transitions\\
Additional study &60 questions, two deployments, four prompts, two budgets & Fixed registered contrasts; 960 assigned outcomes\\
Retrospective fitting &7,140 records; 80/20 response-level split & Association with observed features\\
Other edits &Qwen 15 pairs; Falcon 120 per strategy & Supplied-rationale and prompt revisions\\
Sampling &30 questions, four deployments, three strategies, five draws & Plurality and stop-event checks\\
Candidate audit &100 stratified 30/470 splits per MATH500 deployment & Resampling sensitivity on cached responses\\
\bottomrule
\end{tabular}
\caption{Distinct units and purposes. Repeated strategies and resampling do not add independent questions or deployments.}
\end{table*}

The development tags are \texttt{qwen2.5:7b}, \texttt{gemma3:4b}, \texttt{llama3.2:3b}, \texttt{mistral:7b}, and \texttt{phi4-mini-reasoning:3.8b}. Historical records do not fully retain binary digests, quantization, and serving versions. The additional study captures the current model digests and server configuration before inference. These identify that study, not the earlier generations.

The model creators' cards are linked in the accompanying asset-source index. The additional study records Ollama 0.32.6 and Q4\_K\_M quantization for both Qwen and Phi, with the exact model digests and native templates in \texttt{confirmation/RUNTIME\_MANIFEST\_ANON.json}. The GSM8K and MATH source releases carry MIT licenses; their notices are retained in the artifact. Model weights are not redistributed. The distributed responses and replay code are intended for reproducing the reported comparisons under the respective source terms.

The artifact provides \texttt{data/templates.json}, response-level template bindings, and original request text. CKP uses compact/verbose contrastive examples; MCR-Route combines classification with a development-calibrated routing table; PKR asks for uncertain steps to be identified and expanded. Main-panel KSPAC-Repair includes its multi-stage repair procedure. In MATH500, the archived keys \texttt{kspac\_repair} and \texttt{mcr} instead denote single-generation checking and planning requests, reported as KSPAC-Check and MCR-Instruction. Internal primary, repair, and fallback calls are retained where recorded.

\paragraph{Additional-study prompts.}
The main CoT instruction is ``Solve the problem. Think step by step, then provide the final answer.'' Main CoD asks for a minimal draft with at most five words per reasoning step and no full sentences unless necessary. Both use the historical literal backslash--n encoding around the problem and final-answer request. The full probe asks for ``full step-by-step reasoning before giving the final answer.'' The concise probe asks for ``at most 3 short reasoning lines'' and to ``Avoid all extra explanation.'' These probes use actual newlines. Each requests \texttt{Final answer: <answer>}. Exact bytes are supplied in \texttt{confirmation/templates\_frozen.json} and the frozen request ledger.

The additional study uses temperature zero, seed 20261009, context length 8,192, and output caps 2,048/4,096. Registration is a local pre-run timestamp and hash record, not an external preregistration. Exact-text exclusion against recorded local inputs does not certify absence of semantic overlap or pretraining contamination. Collection amendments concern restart/status handling and terminal interface failures; they preserve original inputs and scoring. Three terminal errors occur in main CoD on Phi. At 2,048 tokens, its mathematical effect is bounded by $[10.0,11.7]$ points under the two possible scores for the missing answer; at 4,096 tokens, the bound is $[8.3,11.7]$. Successful-match effects use the observed endpoint and retain all 60 assignments.

\input{registered_table}
\section{Operation Profiles and All Development Probes}
\label{app:probes}
\begin{table}[t]
\centering\small
\begin{tabular}{llrrrr}
\toprule
Request & Level & Q & G & L & P\\\midrule
Length & Answer-only &47.8&31.1&36.1&55.0\\
 & Concise &54.4&55.0&43.3&56.7\\
 & Full &58.3&51.1&43.9&46.1\\
Structure & Free &39.4&39.4&31.1&53.3\\
 & Simple &56.7&53.9&41.1&52.2\\
 & Multi-stage &64.4&43.9&51.1&52.2\\
Checking & None &48.9&40.0&44.4&55.0\\
 & Self-check &55.0&60.6&43.3&53.3\\
 & Repair &56.7&36.7&35.6&49.4\\
\bottomrule
\end{tabular}
\caption{Marginal accuracy (\%) in the nine-condition design. Q/G/L/P denote Qwen/Gemma/Llama/Phi. Each level pools three conditions on the same 60 questions: 180 responses, not 180 independent questions. These are conditional summaries, not isolated factorial effects.}
\end{table}

\begin{table*}[t]
\centering\small
\begin{tabular}{llrrll}
\toprule
Deployment & Baseline $\to$ target & Correct counts & $\Delta$ (pp) & 95\% interval & Holm $p$\\\midrule
Qwen & Free $\to$ multi-stage &37 $\to$ 42&$+8.3$&$[-1.7,18.3]$&.6797\\
Qwen & Simple $\to$ multi-stage &36 $\to$ 42&$+10.0$&$[-1.7,21.7]$&.5923\\
Gemma & No-check $\to$ self-check &34 $\to$ 38&$+6.7$&$[-8.3,21.7]$&.6797\\
Gemma & Self-check $\to$ repair &38 $\to$ 31&$-11.7$&$[-23.3,0.0]$&.5923\\
Phi & Full $\to$ concise &28 $\to$ 40&$+20.0$&$[8.3,31.7]$&.0293\\
Phi & Full $\to$ answer-only &28 $\to$ 33&$+8.3$&$[-1.7,20.0]$&.6797\\
Phi & Answer-only $\to$ concise &33 $\to$ 40&$+11.7$&$[0.0,23.3]$&.5537\\
\bottomrule
\end{tabular}
\caption{All seven targeted development comparisons, $n=60$ paired questions each. Pointwise intervals and one common Holm family are retained. Qwen/Gemma caps are 1,024 tokens, Phi's is 2,048. Edits can change wording and organization together.}
\end{table*}

\paragraph{Reference-assisted repair records.}
Archived repair probes provide the reference final answer to all three requests: error judgment, localization, and re-answering. The latter two also state that the input solution is wrong. They therefore evaluate responses with external answer information. The counts are retained below with their actual scoring definitions. The judgment task has only inputs labeled wrong, so always answering NO passes its rule; localization uses a text heuristic, not verified first-error annotations. These scores do not estimate independent detection or localization accuracy.

\begin{table}[t]
\centering\small
\setlength{\tabcolsep}{5pt}
\begin{tabular}{lrrrr}
\toprule
Archived pass rule & Qwen & Gemma & Llama & Phi\\\midrule
Output NO &28/32&33/36&50/50&1/37\\
Localization text &32/32&36/36&50/50&32/37\\
Re-answer match &4/32&2/36&10/50&4/37\\
\bottomrule
\end{tabular}
\caption{Reference-assisted repair probes. Phi receives truncated visible traces; other inputs are complete. Localization requires at least 20 characters and specified error/step keywords. Requests are evaluated independently.}
\end{table}

\section{Uniform Answer-Scope Sensitivity}
\label{app:scope}
The frozen answer-matching endpoint uses the complete returned content. To check its dependence on marked thought text, we apply one retrospective rule to all 960 assignments. With no thought markers, use the whole output; with exactly one correctly ordered closed pair, use only the suffix after \texttt{</think>}; with an unclosed or ambiguous marked section, record no observable answer in the specified scope. Interface errors likewise provide no observed match. Apply the archived extractor and matcher to the available scope. The resulting endpoint is an observed outside-thought match, not mathematical correctness of an absent answer or a semantic commitment judgment. It remains separate from the registered analysis.

\begin{table*}[t]
\centering\small
\begin{tabular}{llrrrrr}
\toprule
Deployment & Cap / pair & Base & Target & $R/H$ & $\Delta$ (pp) & Holm $p$\\\midrule
Qwen &2048 / CoT--CoD&48&33&3/18&$-25.0$&.00745\\
Qwen &2048 / full--concise&42&12&1/31&$-50.0$&$1.08\!\times\!10^{-7}$\\
Qwen &4096 / CoT--CoD&46&31&5/20&$-25.0$&.01631\\
Qwen &4096 / full--concise&46&10&0/36&$-60.0$&$2.33\!\times\!10^{-10}$\\
Phi &2048 / CoT--CoD&33&41&12/4&$+13.3$&.23044\\
Phi &2048 / full--concise&32&44&12/0&$+20.0$&.00293\\
Phi &4096 / CoT--CoD&42&47&10/5&$+8.3$&.60352\\
Phi &4096 / full--concise&45&45&3/3&$0.0$&1.00000\\
\bottomrule
\end{tabular}
\caption{Retrospective scope sensitivity, all $n=60$ assignments per cell. Holm is recalculated across these eight sensitivity tests and does not replace Table~\ref{tab:confirmation}. Missing scope is no observed delivered match.}
\end{table*}

For Phi full/concise at 2,048 tokens, 24/16 served outputs have an unclosed or ambiguous marked scope; at 4,096 tokens, the counts are 14/13. Qwen has no unresolved marked scope. The primary frozen-score rescue involving a numeric fallback in unfinished thought text is retained in the original endpoint and excluded by this uniform sensitivity rule. Baseline-only matches are treated by exactly the same rule. Neither an extractor replay nor this structural check constitutes a human semantic review.

\section{Historical Guard Rules and Candidate Audit}
\label{app:guard}
Let $a_m$ be probe matches, $a_{\max}$ the best probe count, and $p_m,e_m,h_m$ the fractions of extraction failures, empty responses, and actual length stops. Let $u_m$ be their union. A candidate is unsupported on the probe when $p_m>0.30$, $e_m>0.05$, or $h_m>0.50$. Otherwise its supported condition is as follows; remaining cases are risky.

The archived term \emph{promotable} denotes a completed audit disposition, not a pre-validation authorization:
\begin{equation}
g_P(m^*)=\mathrm{S},\quad m^*\in\mathrm{NB}(V),\quad m^*\notin\mathrm{F}(V).
\end{equation}
Here NB is the numerical near-best set and F the declared validation reliability set. The other archived dispositions are local-only and unsupported. Their names and thresholds are retained for reproducing the earlier candidate audit; the main-text reporting record does not require this selector.

\begin{table*}[t]
\centering\small
\begin{tabular}{lp{11.5cm}}
\toprule
Candidate & Supported condition\\\midrule
CoT & $u_{\mathrm{CoT}}<0.40$ and the best of the three compression candidates does not exceed CoT by at least 10 points\\
Concise/CoD/KSD & $a_m\ge a_{\max}-1$\\
KSPAC-Check & $a_m\ge\max(a_{\mathrm{CoT}},a_{\mathrm{KSD}})+1$\\
MCR-Instruction & The preceding count condition and $u_m-u_{\mathrm{CoT}}\le0.10$\\
\bottomrule
\end{tabular}
\caption{Development-derived probe rules, replayed with integer counts and denominators. Their provenance does not establish pre-registration before all development observations.}
\end{table*}

Candidates are ordered by match count, event union rate, mean emitted tokens, then the fixed order CoT, Concise CoT, CoD, KSD, KSPAC-Check, MCR-Instruction. Guard retains supported candidates, falling back to the first candidate if none is supported without changing its label. Validation near-best uses a $\max(2,\lceil0.03n\rceil)$ match-count tolerance. The validation reliability set flags an event union rate above 5\%, an accuracy gap of at least 15 points from the best candidate, or accuracy at most 20\%. These are declared reporting criteria, not statistical risk guarantees.

Each MATH500 replay samples ten probe questions per difficulty stratum, with 30 probe and 470 validation questions. There are 100 repetitions per deployment, seed 20260726. A hit means at least one retained candidate belongs to the target set, including tied maxima; it is not recall of the entire set. Aggregate values first average within deployment and then across the three deployments.

\begin{table}[t]
\centering\small
\setlength{\tabcolsep}{5pt}
\begin{tabular}{lrrr}
\toprule
Rule & Size & Near-best & Highest point\\\midrule
Direct top-1 &1.00&67.0&34.7\\
Direct top-2 &2.00&86.0&54.7\\
Direct top-3 &3.00&96.0&76.0\\
Probe top-band &5.55&100.0&100.0\\
Guard &2.18&81.0&72.3\\
Direct, matched size &2.18&86.0&66.3\\
\bottomrule
\end{tabular}
\caption{Candidate size and hit rates (\%) across cached splits. Direct top-2 plus the same hard interface filter is unchanged at 86.0/54.7.}
\end{table}

\paragraph{Retention and evaluation budget.}
The 100 stratified 30/470 splits per deployment overlap and measure sensitivity to sampled probe questions. Highest-point hit rates are 54.7\% for Direct top-2 and 72.3\% for Guard; matching Direct's candidate count to Guard gives 86.0\% near-best and 66.3\% highest-point hits. The table reports a hit in the target set, rather than recall of every member. On a new deployment, evaluating the full six-candidate reference pool costs $6\times30=180$ probe evaluations and $6\times470=2,820$ validation evaluations per split, plus any operation probes. Multi-stage strategies incur their internal calls as well. Cached replay is not new inference, and retaining fewer candidates does not avoid the cost of evaluating the reference pool.

The archived probe top-band comparator tests the point winner against its five alternatives; it retains that selection-dependent implementation as a candidate-retention reference. Main MATH500 inference instead uses all 15 pairs. Multiplying all interface thresholds by 0.75 or 1.25 changes none of 1,800 probe labels in this panel. Changing the compression tolerance from one answer to zero/two changes 103/105 labels and mean size to 1.83/2.53. Near-best tolerances of 2\%/4\% give Guard hit rates 80.7\%/83.3\%, versus 81.0\% at 3\%. These retrospective checks characterize the rules without selecting new thresholds from their outcomes.

\section{Sampling and Reproduction}
\label{app:sampling}
The sampling block draws five answers per strategy for CoT, CoD, and KSD on 30 stratified questions in four deployments, using temperature 0.7, top-$p$ 0.95, and seeds 101--105. It chooses the most frequent extracted answer string; ties prefer nonempty strings and then lexical order. Mathematical equivalence is not used to merge strings before voting. Of 360 voting units, 166 lack a three-of-five majority and 125 have a tied maximum. We therefore describe this as plurality aggregation.

The largest plurality accuracies are Qwen KSD 83.3\%, Gemma KSD 80.0\%, Llama CoT 73.3\%, and Phi CoT 6.7\%. Across 1,800 generations, there are 45 actual length stops, including 30 on Phi. The settings and ties accompany these results; they do not imply that sampling is generally ineffective.

The evidence package contains response and template indexes, scoring outputs, all comparison families, transitions, threshold replays, and figure inputs. Original extraction and matching code is preserved. The 14,400-response replay has zero correctness-label differences. Additional repeatability records and single-author AI-assisted annotations have separate indexes (Appendices~\ref{app:repeatability} and~\ref{app:author_annotation}); they neither replace original scores nor certify all rationales. Replaying extraction, matching references, and judging answer commitments remain distinct checks.

\section{Paired Effects Under the Primary Scoring}
The following two tables give the complete CoT-relative comparisons used to interpret answer transitions in the two main data blocks.
\input{paired_tables}

%% file: registered_table.tex
\subsection{Complete Registered Contrasts}\label{app:registered_all}
Table~\ref{tab:confirmation} reports all eight registered comparisons on the additional question set, retaining the frozen endpoint and common correction family.
\begin{table*}[t]
\centering\small
\begin{tabular}{llrrrrrl}
\toprule
Deployment & Cap / pair & Base & Target & $R/H$ & $\Delta$ (pp) & Holm $p$ & 95\% paired interval\\
\midrule
Qwen & 2048 / CoT--CoD &48&33&3/18&$-25.0$&.00894&$[-38.3,-11.7]$\\
Qwen & 2048 / full--concise &42&12&1/31&$-50.0$&$1.08\!\times\!10^{-7}$&$[-63.3,-36.7]$\\
Qwen & 4096 / CoT--CoD &46&31&5/20&$-25.0$&.02039&$[-40.0,-10.0]$\\
Qwen & 4096 / full--concise &46&10&0/36&$-60.0$&$2.33\!\times\!10^{-10}$&$[-71.7,-48.3]$\\
Phi & 2048 / CoT--CoD &37&43&8/2&$+10.0$&.32813&$[0.0,20.0]$\\
Phi & 2048 / full--concise &37&45&8/0&$+13.3$&.03125&$[5.0,21.7]$\\
Phi & 4096 / CoT--CoD &45&50&7/2&$+8.3$&.35938&$[-1.7,18.3]$\\
Phi & 4096 / full--concise &46&46&3/3&$0.0$&1.00000&$[-8.3,8.3]$\\
\bottomrule
\end{tabular}
\caption{All eight registered contrasts on 60 additional questions per cell. Counts use the frozen answer-matching endpoint; intervals are pointwise. Phi CoD has one terminal interface failure at 2,048 tokens and two at 4,096; these yield no successful match, with unknown mathematical correctness. Full/concise are probe templates, not the main CoT/CoD templates.}
\label{tab:confirmation}
\end{table*}

%% file: paired_tables.tex
\begin{table*}[t]
\centering\small
\begin{tabular}{llrrrrrl}
\toprule
Deployment & Target & $n$ & Base & Target & $R/H$ & $\Delta$ (pp) & 95\% interval\\\midrule
Gemma & CoD & 120 & 84 & 61 & 4/27 & $-19.2$ & $[-27.5,-10.8]$ \\
Gemma & KSD & 120 & 84 & 87 & 11/8 & $+2.5$ & $[-4.2,10.0]$ \\
Gemma & Concise & 120 & 84 & 79 & 7/12 & $-4.2$ & $[-11.7,2.5]$ \\
Llama & CoD & 120 & 69 & 60 & 15/24 & $-7.5$ & $[-17.5,2.5]$ \\
Llama & KSD & 120 & 69 & 84 & 25/10 & $+12.5$ & $[3.3,21.7]$ \\
Llama & Concise & 120 & 69 & 69 & 16/16 & $+0.0$ & $[-9.2,9.2]$ \\
Mistral & CoD & 120 & 12 & 11 & 6/7 & $-0.8$ & $[-6.7,5.0]$ \\
Mistral & KSD & 120 & 12 & 7 & 4/9 & $-4.2$ & $[-10.0,1.7]$ \\
Mistral & Concise & 120 & 12 & 6 & 4/10 & $-5.0$ & $[-10.8,0.8]$ \\
Phi & CoD & 120 & 83 & 98 & 20/5 & $+12.5$ & $[5.0,20.8]$ \\
Phi & KSD & 120 & 83 & 82 & 5/6 & $-0.8$ & $[-5.8,5.0]$ \\
Phi & Concise & 120 & 83 & 93 & 16/6 & $+8.3$ & $[0.8,15.8]$ \\
Qwen & CoD & 120 & 88 & 56 & 5/37 & $-26.7$ & $[-35.8,-17.5]$ \\
Qwen & KSD & 120 & 88 & 88 & 10/10 & $+0.0$ & $[-7.5,7.5]$ \\
Qwen & Concise & 120 & 88 & 87 & 13/14 & $-0.8$ & $[-9.2,7.5]$ \\
\bottomrule
\end{tabular}
\caption{Development paired effects relative to CoT. Intervals are pointwise; full comparison families determine statistical top bands.}
\end{table*}
\begin{table*}[t]
\centering\small
\begin{tabular}{llrrrrrl}
\toprule
Deployment & Target & $n$ & Base & Target & $R/H$ & $\Delta$ (pp) & 95\% interval\\\midrule
Gemma & CoD & 500 & 354 & 253 & 20/121 & $-20.2$ & $[-24.6,-16.0]$ \\
Gemma & KSD & 500 & 354 & 340 & 27/41 & $-2.8$ & $[-6.0,0.4]$ \\
Gemma & Concise & 500 & 354 & 343 & 30/41 & $-2.2$ & $[-5.6,1.0]$ \\
Gemma & Check & 500 & 354 & 374 & 41/21 & $+4.0$ & $[1.0,7.0]$ \\
Gemma & Instruction & 500 & 354 & 347 & 33/40 & $-1.4$ & $[-4.6,2.0]$ \\
Llama & CoD & 500 & 210 & 123 & 33/120 & $-17.4$ & $[-22.0,-12.8]$ \\
Llama & KSD & 500 & 210 & 189 & 48/69 & $-4.2$ & $[-8.4,0.0]$ \\
Llama & Concise & 500 & 210 & 173 & 52/89 & $-7.4$ & $[-12.0,-2.8]$ \\
Llama & Check & 500 & 210 & 207 & 65/68 & $-0.6$ & $[-5.2,4.0]$ \\
Llama & Instruction & 500 & 210 & 201 & 54/63 & $-1.8$ & $[-6.2,2.4]$ \\
Qwen & CoD & 500 & 363 & 197 & 15/181 & $-33.2$ & $[-38.0,-28.6]$ \\
Qwen & KSD & 500 & 363 & 339 & 33/57 & $-4.8$ & $[-8.4,-1.0]$ \\
Qwen & Concise & 500 & 363 & 349 & 36/50 & $-2.8$ & $[-6.4,0.8]$ \\
Qwen & Check & 500 & 363 & 357 & 34/40 & $-1.2$ & $[-4.6,2.2]$ \\
Qwen & Instruction & 500 & 363 & 347 & 22/38 & $-3.2$ & $[-6.2,-0.2]$ \\
\bottomrule
\end{tabular}
\caption{MATH500 paired effects relative to CoT. Intervals are pointwise; full comparison families determine statistical top bands.}
\end{table*}

%% file: theory_appendix.tex
\section{Paired Execution Model: Assumptions and Derivations}
\label{app:theory}

\paragraph{Objects and assumptions.}
The input $X$ includes the question and the evaluator's reference information, which is not thereby supplied to the generator. Records contain text, execution events, and a terminal error symbol when no response is served. Strategies include their invocation and fallback rules. The response marginals of $K_e(\cdot\mid X)$ must match the declared collection procedures. Independence between the two responses is unnecessary. With stochastic generation, rescue and harm probabilities depend on the pairing protocol, although their difference depends only on marginal accuracies.

The evaluator is a fixed, measurable binary function of the input and response. We treat answer matching as the operational endpoint; it is not a definition of the semantic validity of a rationale. A terminal interface failure receives zero for observed successful matching. Its unknown mathematical answer remains missing. Execution categories form a finite, exhaustive partition with the same meaning in each compared configuration. Our stop-pair partition has nine possible categories, including failure categories; unused categories have zero mass. A missing stop field would require its own category rather than silent exclusion.

For the deployment comparison, strategies, $Q$, and the evaluator are held fixed. Changing the budget is part of changing the deployment. If the task distribution also changes, the algebra still holds for the resulting joint laws, but the difference then includes task shift and cannot be attributed to deployment alone. A scorer change is handled separately below. No structural equation for a latent ``reasoning capacity'' is assumed.

\paragraph{Paired decomposition.}
For binary labels, $D=Y_1-Y_0$ equals one on a rescue, minus one on a harm, and zero otherwise. Hence
\begin{equation}
\delta_e=\Pr_e(Y_0=0,Y_1=1)-\Pr_e(Y_0=1,Y_1=0).
\end{equation}
Partitioning these probabilities over $Z$ proves Equation~\ref{eq:joint}. On a finite ledger, let $n_z,R_z,H_z$ be category counts. Its exact empirical version is
\begin{equation}
\widehat\delta_e=\sum_{z:n_z>0}\frac{n_z}{n}\frac{R_z-H_z}{n_z}.
\end{equation}
Equal accuracy therefore constrains only the difference of rescues and harms. It does not require identical correct-answer sets. The Qwen CoT--KSD comparison in Figure~\ref{fig:paired} realizes this distinction with ten transitions in each direction.

\paragraph{Proof of the transfer bound.}
Abbreviate $w_e(z),w_{e'}(z),\mu_e(z),\mu_{e'}(z)$ by $w_z,w'_z,\mu_z,\mu'_z$ and put $v_z=\min(w_z,w'_z)$. A zero-mass conditional mean may take any value in $[-1,1]$, since its coefficient is zero. Separate common and excess mass:
\begin{equation}
\begin{split}
\delta_{e'}-\delta_e={}&\sum_z v_z(\mu'_z-\mu_z)\\
&+\sum_z(w'_z-v_z)\mu'_z\\
&-\sum_z(w_z-v_z)\mu_z.
\end{split}
\end{equation}
Because $|\mu_z|,|\mu'_z|\le1$ and both weight vectors sum to one, each excess-mass sum has total weight $1-\sum_z v_z=\mathrm{TV}(w,w')$. Taking absolute values proves Equation~\ref{eq:transfer}. Its right-hand side $B$ is at most two: $|\mu'_z-\mu_z|\le2$ gives $B\le2\mathrm{TV}(w,w')+2\sum_z v_z=2$. The bound can nevertheless be loose for a particular comparison. If $\delta_e>B$, then $\delta_{e'}\ge\delta_e-B>0$. This is a sufficient condition involving both deployments' joint distributions, not a prediction rule using the source deployment alone.

\paragraph{Why stopping alone is insufficient.}
Consider one fixed question and deterministic response records whose stop category is normal-to-normal. In deployment $e$, the baseline answer is wrong and the target correct, giving $w_e(z)=1$ and $\mu_e(z)=1$. In $e'$, let the baseline be correct and the target wrong, with the same stop category. Then $w_{e'}=w_e$ but $\mu_{e'}(z)=-1$, and the effect changes from $+1$ to $-1$. This construction satisfies the model with a common evaluator and input distribution. Thus even the full stop-pair distribution cannot identify the direction of task benefit. Conditioning on a post-generation event also does not amount to intervening on that event.

\paragraph{Scorer sensitivity.}
Hold the response law fixed and apply two deterministic evaluators, $s$ and $\widetilde s$. Let $\epsilon_j=\Pr(s(X,O_j)\ne\widetilde s(X,O_j))$. Subtracting their paired effects and using the triangle inequality gives
\begin{equation}
|\widetilde\delta_e-\delta_e|\le\epsilon_0+\epsilon_1.
\label{eq:scorer}
\end{equation}
Indeed, the absolute change in each binary score is exactly its disagreement indicator. This bound requires checking both arms. For Phi full--concise at 2,048 tokens, the uniform scope check changes five baseline matches and one target match to non-matches, without adding matches. Thus $\epsilon_0=5/60$, $\epsilon_1=1/60$, and the observed effect change, $4/60$, is at most $6/60$. Appendix~\ref{app:scope} applies that check to every registered condition; the frozen primary result is not replaced.

\paragraph{Empirical and population readings.}
All decompositions apply exactly to the observed paired ledger without sampling assumptions. Reading its estimates as expectations for future items additionally requires a relevant sampling design and stability of the declared response procedures. Repeated runs on one item or overlapping resampling splits are not independent new items. The paired bootstrap preserves each observed response pair \citep{efron1979}; its intervals quantify resampling variability and do not cover deployment shift. For independent, identically distributed response pairs, the exact conditional McNemar test conditions on the number of discordant pairs and uses a binomial probability of one half under marginal equality \citep{mcnemar1947,fagerland2013}. The fixed question strata and development selection therefore accompany the interpretation of those tests. Holm adjustment applies to the stated comparison family \citep{holm1979}, not to an undisclosed search over candidate prompts.

The supplementary checks compare the derivations with finite distributions, boundary cases, and the reported empirical decomposition.

%% file: followup_tables.tex
\newpage
\section{Additional Retrospective Controls}

\label{app:controls}

These analyses reuse the archived responses. They are additional retrospective controls, not new model generations or registered primary tests.

\paragraph{Information-matched fitting.}

All rows use the original 5,712/1,428 record split (seed 42), balanced logistic regression, and identical targets. Length is the log of recorded output tokens. Control features are the four archived compression, structure, verification, and repair descriptors. The split can share questions and deployments; AUC differences have not been significance-tested.

\par\medskip\noindent\begin{minipage}{\columnwidth}\centering\small

\begin{tabular}{lrr}

\toprule

Features & Correctness & Interface event\\\midrule

Length & 0.7029 & 0.8336\\

Deploy. + length & 0.7940 & 0.9131\\

Method + deploy. + len. & 0.8173 & 0.9433\\

Deploy. + control & 0.7040 & 0.8920\\

Enriched & 0.8021 & 0.9432\\

\bottomrule\end{tabular}

\captionof{table}{Added same-split AUC controls. Enriched is the original descriptor set in Figure~\ref{fig:original}. Output length is observed after generation, not a pre-inference measurement.}\label{tab:auc_control}\end{minipage}\par\medskip

\input{historical_fit}

\paragraph{Execution and answer scope.}

Table~\ref{tab:execution_metrics} retains the same full/concise responses as Table~\ref{tab:confirmation}. The scope endpoint applies one structural rule to both arms. It does not certify a final semantic commitment.

\begin{table*}[t]\centering\small

\begin{tabular}{llrrrrr}

\toprule

Deployment / cap & Request & Matches & Scope matches & Median tokens & Length stops & Unresolved scope\\\midrule

Qwen / 2048 & full & 42 & 42 & 426.5 & 0 & 0\\

Qwen / 2048 & concise & 12 & 12 & 4.5 & 0 & 0\\

Qwen / 4096 & full & 46 & 46 & 422.0 & 0 & 0\\

Qwen / 4096 & concise & 10 & 10 & 4.0 & 0 & 0\\

Phi / 2048 & full & 37 & 32 & 1864.0 & 26 & 24\\

Phi / 2048 & concise & 45 & 44 & 936.0 & 16 & 16\\

Phi / 4096 & full & 46 & 45 & 1858.0 & 15 & 14\\

Phi / 4096 & concise & 46 & 45 & 940.5 & 12 & 13\\

\bottomrule\end{tabular}

\caption{Observed execution on the 60 additional questions in each cell. All 480 full/concise responses were served; the three interface failures occur in a different arm.}\label{tab:execution_metrics}\end{table*}

\begin{table*}[t]\centering\small

\begin{tabular}{llrrrrr}

\toprule

Deployment / cap & Stop pair & $n$ & Rescue & Harm & Both & Neither\\\midrule

Qwen / 2048 & normal$\to$normal & 60 & 1 & 31 & 11 & 17\\

Qwen / 4096 & normal$\to$normal & 60 & 0 & 36 & 10 & 14\\

Phi / 2048 & length$\to$length & 15 & 0 & 0 & 0 & 15\\

Phi / 2048 & length$\to$normal & 11 & 7 & 0 & 4 & 0\\

Phi / 2048 & normal$\to$length & 1 & 1 & 0 & 0 & 0\\

Phi / 2048 & normal$\to$normal & 33 & 0 & 0 & 33 & 0\\

Phi / 4096 & length$\to$length & 10 & 0 & 1 & 0 & 9\\

Phi / 4096 & length$\to$normal & 5 & 3 & 0 & 1 & 1\\

Phi / 4096 & normal$\to$length & 2 & 0 & 2 & 0 & 0\\

Phi / 4096 & normal$\to$normal & 43 & 0 & 0 & 42 & 1\\

\bottomrule\end{tabular}

\caption{Complete observed stop-pair categories for full-to-concise. Missing categories have zero observations. Row cells partition each 60-question comparison.}\label{tab:complete_joint}\end{table*}

At 4,096 tokens on Phi, three rescues in length-to-normal pairs are offset by three harms in pairs ending at the length cap. On Qwen, all pairs stop normally despite the large loss in matches. Completion and correctness therefore remain distinct measurements.

\paragraph{Stratified sensitivity.}

A separate retrospective bootstrap resamples 20 questions within each of the three strata, using NumPy default\_rng(20261011) and 20,000 draws. For Phi at 2,048 tokens, its pointwise interval is [5.0, 23.3] pp, with stratum rescue/harm counts 2/0, 3/0, and 3/0. This sensitivity does not replace the registered pooled interval [5.0, 21.7] pp or its Holm family. It likewise does not estimate repeated-generation variation.

\section{Additional-Sample Selection Replay}

\label{app:selection}

The source snapshot contains GSM8K test and seven MATH test subjects. Keeping MATH rows with a final boxed reference gives 6,319 candidates. A Python random.Random(20261009) generator shuffles each stratum in the order GSM8K, MATH levels 1--3, MATH levels 4--5. NFKC-normalized, case-folded alphanumeric question text identifies duplicates. The draw skips duplicates and recorded archived-input exclusions, takes the first 20 eligible items in each stratum, and shuffles the combined 60 using the same generator.

The encountered exclusion IDs are \texttt{math\_test\_number\_theory\_0361} and \texttt{math\_test\_precalculus\_0442}. The artifact supplies the eight source files with dataset revisions and SHA256 hashes, the selected ledger, and \texttt{scripts/replay\_selection.py}. The replay checks all selected fields and their order conditional on these exclusions. It does not retrospectively authenticate all historical archive scans or the pre-run version of the selector source. No generated-answer score enters the draw.

The study estimates a balanced three-stratum comparison, rather than a prevalence-weighted score for the original benchmark mixture. The fixed primary analysis remains intact; the additional within-stratum sensitivity is reported separately.

%% file: historical_fit.tex
\twocolumn[{
\subsection{Historical Fitting and Prompt Edits}\label{app:historical_fit}
\centering
\includegraphics[width=\textwidth]{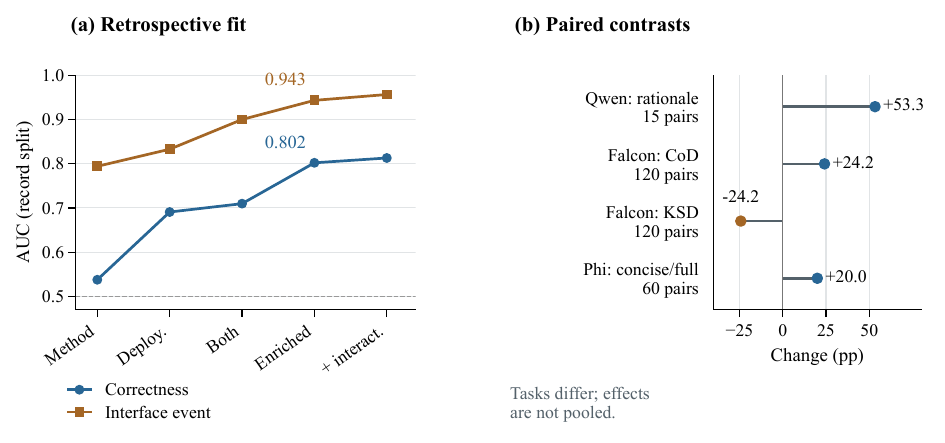}
\captionof{figure}{Retrospective fit and concrete prompt changes. (a) Method, Deploy., and Both use identifiers; Enriched combines deployment identity, control descriptions, and observed output length, followed by interactions. Targets are answer correctness and interface events. This is a record-level split, not unseen-deployment prediction. (b) Qwen compares supplied rationale lengths (15 pairs); Falcon uses separate CoD/KSD prompt revisions (120 pairs each); Phi uses the 60-question development full-to-concise contrast. Effects retain their own experimental settings and denominators.}
\label{fig:original}
\par\medskip
}]
Figure~\ref{fig:original} relates two further analyses of execution. On 7,140 responses split 80/20 by record, enriched features achieve AUC 0.802 for correctness and 0.943 for interface events. An added same-split control using method, deployment, and observed output length reaches 0.817 and 0.943, respectively (Appendix~\ref{app:controls}). Control descriptions thus do not improve these point estimates over the simpler descriptor baseline. Questions and deployments can occur in both partitions; these are retrospective fits, not unseen-deployment predictions. The interface-event target is distinct from Guard's validation reliability criterion.

The paired changes in the second panel show why the actual edit matters. Increasing supplied rationale length on Qwen changes 7/15 to 15/15 matches. Separate Falcon revisions change CoD from 3/120 to 32/120 and KSD from 64/120 to 35/120. They are method-specific prompt edits, not one common backend intervention. Phi's point is the development contrast in Table~\ref{tab:joint}.

%% file: repeat_annotation_tables.tex
\section{Repeated Collection Under Fixed Settings}

\label{app:repeatability}

We collected the full/concise pair twice more on the same 60 questions at both caps and deployments: 960 new responses in 16 cells. Seed 20261009, temperature zero, the native templates, context length 8,192, model digests, and Ollama version 0.32.6 were retained. Each collection is reported separately; repeated responses are not additional independent questions.

The eight new full-to-concise tests form a separate Holm family. Table~\ref{tab:repeat_all} reports every contrast. The original four-prompt study and its eight-way correction are unchanged.

\begin{table*}[t]\centering\small

\begin{tabular}{llrrrrrr}

\toprule

Deployment & Cap & Collection & Base & Target & Rescue/Harm & Change (pp) & Holm $p$\\\midrule

Phi & 2048 & 1 & 35 & 46 & 14/3 & $+18.3$ & $0.0381775$\\

Phi & 2048 & 2 & 34 & 48 & 14/0 & $+23.3$ & $0.000488281$\\

Phi & 4096 & 1 & 41 & 48 & 10/3 & $+11.7$ & $0.18457$\\

Phi & 4096 & 2 & 46 & 51 & 6/1 & $+8.3$ & $0.18457$\\

Qwen & 2048 & 1 & 42 & 12 & 2/32 & $-50.0$ & $3.469\!\times\!10^{-7}$\\

Qwen & 2048 & 2 & 43 & 12 & 1/32 & $-51.7$ & $4.750\!\times\!10^{-8}$\\

Qwen & 4096 & 1 & 44 & 10 & 0/34 & $-56.7$ & $9.313\!\times\!10^{-10}$\\

Qwen & 4096 & 2 & 42 & 10 & 1/33 & $-53.3$ & $2.852\!\times\!10^{-8}$\\

\bottomrule\end{tabular}

\caption{All eight new repeated-collection contrasts; each uses the same 60 paired questions and original answer-matching endpoint. No new independent sample is claimed.}\label{tab:repeat_all}\end{table*}

Phi's gain repeats at 2,048 tokens and passes correction in both collections. At 4,096, the estimates are positive but do not pass correction. Together with the original zero net difference, these observations show variation at the larger cap rather than a consistent disappearance of the gain. Qwen loses matches in every collection.

\begin{table*}[t]\centering\small

\begin{tabular}{llrrrrrr}

\toprule

Deploy. / cap & Run & Label eq. & Stop eq. & Text eq. & Output tok. & Input tok. & Time (s)\\\midrule

Phi / 2048 & 1 & 103/120 & 107/120 & 0/120 & 163027 & 12906 & 2671.0\\

Phi / 2048 & 2 & 110/120 & 112/120 & 0/120 & 157502 & 12906 & 2556.5\\

Phi / 4096 & 1 & 109/120 & 110/120 & 0/120 & 236087 & 12906 & 4396.3\\

Phi / 4096 & 2 & 105/120 & 107/120 & 0/120 & 221535 & 12906 & 3589.0\\

Qwen / 2048 & 1 & 114/120 & 120/120 & 55/120 & 36508 & 13000 & 816.7\\

Qwen / 2048 & 2 & 115/120 & 120/120 & 60/120 & 36371 & 13000 & 795.3\\

Qwen / 4096 & 1 & 114/120 & 120/120 & 58/120 & 34998 & 13000 & 766.6\\

Qwen / 4096 & 2 & 112/120 & 120/120 & 57/120 & 36882 & 13000 & 805.3\\

\bottomrule\end{tabular}

\caption{Agreement with the original collection and observed cost, pooling the 60 full and 60 concise responses only for descriptive agreement counts. Input-token counts need not equal uncached computation.}\label{tab:repeat_cost}\end{table*}

Relative to the original collection, 882 of the 960 new outputs have the same correctness label, 916 have the same stop state, and 230 have exactly the same text. Generation therefore varies despite fixed sampling settings, although these records do not identify its source. The artifact verifies frozen requests, input and runtime-source hashes, and extracted-answer matches while preserving the original records.

\section{Author-Verified Answer-Commitment Annotation}

\label{app:author_annotation}

An additional retrospective review covered all 120 responses in the original Phi full/concise comparison at 2,048 tokens. A single author verified AI-assisted initial annotations. Response order was randomized and method names, model tags, and automatic scores were hidden in the review kit. The author states that the mapping was not inspected before annotation; response style can nevertheless reveal a condition. This is an author-verified single review, not an independent double review.

The rubric distinguishes final-answer commitment, reference equivalence, rationale validity, and uncertainty. An unclosed thought marker alone does not settle whether a response commits to an answer. Likewise, absence of a final answer is not a judgment of mathematical error. The artifact preserves every label and original note; the English manuscript reports their aggregate results.

The exact original English rubric, Chinese operational instructions, blinded response view and blank forms are preserved in \texttt{answer\_annotations/original\_review\_kit}. Its then-pending wording is historical; the received author-verified labels are retained separately. The rubric asks for a readable, unambiguous commitment rather than an intermediate candidate, and does not decide commitment from a thought marker alone. Boundary illustrations include B002, accepted despite an unclosed marker after repeated conclusions; B076, where the author inferred a count from an exhaustive list; and B060/B113, unresolved because later candidates conflict. These are descriptions of the received judgments, not new rules retrospectively assigned to the original review. No labels were recoded in this revision.

\begin{table}[t]\centering\small

\begin{tabular}{lrrrr}

\toprule

Request & Equiv. & Not equiv. & No final & Uncertain\\\midrule

Full & 44 & 1 & 13 & 2\\

Concise & 48 & 0 & 10 & 2\\

\bottomrule\end{tabular}

\caption{All 60 responses per request under the additional commitment rubric. Uncertain outcomes remain unresolved rather than being labeled mathematical errors.}\label{tab:author_labels}\end{table}

Confirmed equivalent commitments give 44 versus 48 answers, with six rescues, two harms, and a +6.7 pp net gain. The exploratory paired test gives $p=0.2891$. Across all 16 binary completions of the four unresolved labels, the gain ranges from +3.3 to +10.0 pp, with no nominal $p$ below 0.05. This range describes the unresolved labels, not sampling uncertainty or annotation error.

There are 11 resolved disagreements with frozen answer matching and four unresolved cases. Among rationale labels, 118 are not checked and two are marked invalid despite equivalent final answers. Neither final-answer equivalence nor the original score replay validates the underlying derivation. The annotation is separate from all frozen test families and does not adjudicate the new repeated outputs.

%% file: consolidation_appendix.tex
\section{Repeated-Collection and Path Summaries}\label{app:consolidation}
\raggedbottom

We summarize the completed records retrospectively. The dated specification and code are archived separately from the original registered analyses. These summaries add no generations or primary tests.

For each model and budget, form a $60\times3$ matrix of concise-minus-full labels from the three chat collections. Average within question and then across the balanced strata. Question-cluster resampling retains all three collection observations. The joint sensitivity additionally resamples three collection indices globally within a draw, shared across the two budgets. Both use 20,000 draws and numpy default\_rng(20261012). This preserves within-question and cross-budget pairing; three observed collections do not identify the distribution of arbitrary future serving changes.

Table~\ref{tab:cluster} reports both pointwise summaries. The paired 2,048-minus-4,096 effect contrasts are:

Phi: $+11.7$ pp, question-cluster interval $[5.0,18.9]$, joint interval $[1.7,21.1]$.

Qwen: $+6.1$ pp, question-cluster interval $[1.1,11.7]$, joint interval $[0.0,13.9]$.

Both contrasts are descriptive and add no primary significance decisions.

\input{path_table}

For Phi/2,048, both A and W rescue eleven questions on two-call target paths and two on one-call paths. A has one two-call harm; W has none. The comparisons also involve regenerated first-stage responses and the warning/budget procedure, so they do not isolate the effects of continuation, thought closure, or budget reservation.

%% file: path_table.tex
\begin{table}[t]\centering\small
\setlength{\tabcolsep}{3pt}
\begin{tabular}{lrrrrrrr}
\toprule
Cell & Arm & Calls & $n$ & $R$ & $H$ & $B$ & $N$\\\midrule
Phi / 2048 & A & 1 & 34 & 2 & 0 & 32 & 0\\
Phi / 2048 & A & 2 & 26 & 11 & 1 & 5 & 9\\
Phi / 2048 & W & 1 & 31 & 2 & 0 & 29 & 0\\
Phi / 2048 & W & 2 & 29 & 11 & 0 & 9 & 9\\
Phi / 4096 & A & 1 & 46 & 0 & 0 & 45 & 1\\
Phi / 4096 & A & 2 & 14 & 9 & 0 & 1 & 4\\
Phi / 4096 & W & 1 & 42 & 0 & 0 & 41 & 1\\
Phi / 4096 & W & 2 & 18 & 8 & 0 & 5 & 5\\
Qwen / 2048 & A & 1 & 60 & 1 & 2 & 41 & 16\\
Qwen / 2048 & A & 2 & 0 & 0 & 0 & 0 & 0\\
Qwen / 2048 & W & 1 & 59 & 7 & 2 & 40 & 10\\
Qwen / 2048 & W & 2 & 1 & 0 & 1 & 0 & 0\\
Qwen / 4096 & A & 1 & 60 & 0 & 2 & 42 & 16\\
Qwen / 4096 & A & 2 & 0 & 0 & 0 & 0 & 0\\
Qwen / 4096 & W & 1 & 59 & 7 & 4 & 39 & 9\\
Qwen / 4096 & W & 2 & 1 & 0 & 1 & 0 & 0\\
\bottomrule\end{tabular}
\caption{Full-to-A/W outcomes by realized target call path. R/H: rescue/harm; B/N: both/neither correct. Each two-row partition sums to its 60 pairs and registered R/H. Triggering is an observed event, not an assigned causal treatment.}\label{tab:paths}
\end{table}

%% file: answer_phase_tables.tex
\section{Equal-Generated-Cap Answer-Phase Comparison}

\label{app:answer_phase}

We compare full CoT, the concise request, and two answer-phase controls on the original 60 questions in each deployment/cap. This block reuses the questions rather than adding a held-out sample. The controls adapt the early-answer approach of \citet{sun-etal-2025-empirical} as existing-method baselines for the compression interpretation.

All four arms use native prefixes through the same raw generation interface, with temperature zero, seed 20261009, context 8,192, and the recorded model digests and Ollama 0.32.6. The two controls reserve 128 tokens from cap $B$, generate at most $B-128$ initially, and continue the same assistant prefix only on a length stop. A normal stop uses the initial response. The warned control appends: \emph{If [ANSWER NOW] appears, stop the intermediate work and immediately give your best final answer.}

The controller closes an unmatched thought marker, when present, then supplies [ANSWER NOW] and the fixed Final answer: slot. These are input injections, not generated answer text. The primary score uses this slot plus the generated continuation, or the original response when no continuation occurs. Prespecified sensitivities score the generated suffix alone and the whole returned trace. All 960 matching labels agree across the three endpoints; their three separately corrected 24-test families therefore have identical numerical results. This does not certify semantic answer commitment or intermediate reasoning.

\newsavebox{\comparisonledger}
\begin{lrbox}{\comparisonledger}\begin{minipage}{\textwidth}\centering\small

\begin{tabular}{llrrrrrr}

\toprule

Deployment / cap & Edit & Base & Target & R/H & Change (pp) & 95\% interval & Holm $p$\\\midrule

Phi / 2048 & F $\to$ C & 38 & 41 & 6/3 & $+5.0$ & $[-5.0,15.0]$ & $1$\\

Phi / 2048 & F $\to$ A & 38 & 50 & 13/1 & $+20.0$ & $[10.0,31.7]$ & $0.031128$\\

Phi / 2048 & F $\to$ W & 38 & 51 & 13/0 & $+21.7$ & $[11.7,31.7]$ & $0.0043945$\\

Phi / 2048 & C $\to$ A & 41 & 50 & 11/2 & $+15.0$ & $[5.0,25.0]$ & $0.29199$\\

Phi / 2048 & C $\to$ W & 41 & 51 & 11/1 & $+16.7$ & $[6.7,26.7]$ & $0.095215$\\

Phi / 2048 & A $\to$ W & 50 & 51 & 3/2 & $+1.7$ & $[-5.0,8.3]$ & $1$\\

Phi / 4096 & F $\to$ C & 46 & 50 & 4/0 & $+6.7$ & $[1.7,13.3]$ & $1$\\

Phi / 4096 & F $\to$ A & 46 & 55 & 9/0 & $+15.0$ & $[6.7,23.3]$ & $0.0625$\\

Phi / 4096 & F $\to$ W & 46 & 54 & 8/0 & $+13.3$ & $[6.7,21.7]$ & $0.10938$\\

Phi / 4096 & C $\to$ A & 50 & 55 & 5/0 & $+8.3$ & $[1.7,15.0]$ & $0.75$\\

Phi / 4096 & C $\to$ W & 50 & 54 & 6/2 & $+6.7$ & $[-1.7,15.0]$ & $1$\\

Phi / 4096 & A $\to$ W & 55 & 54 & 1/2 & $-1.7$ & $[-6.7,3.3]$ & $1$\\

Qwen / 2048 & F $\to$ C & 43 & 12 & 2/33 & $-51.7$ & $[-65.0,-36.7]$ & $7.713\times10^{-7}$\\

Qwen / 2048 & F $\to$ A & 43 & 42 & 1/2 & $-1.7$ & $[-6.7,3.3]$ & $1$\\

Qwen / 2048 & F $\to$ W & 43 & 47 & 7/3 & $+6.7$ & $[-3.3,16.7]$ & $1$\\

Qwen / 2048 & C $\to$ A & 12 & 42 & 32/2 & $+50.0$ & $[35.0,63.3]$ & $1.388\times10^{-6}$\\

Qwen / 2048 & C $\to$ W & 12 & 47 & 36/1 & $+58.3$ & $[45.0,70.0]$ & $1.327\times10^{-8}$\\

Qwen / 2048 & A $\to$ W & 42 & 47 & 9/4 & $+8.3$ & $[-3.3,20.0]$ & $1$\\

Qwen / 4096 & F $\to$ C & 44 & 13 & 1/32 & $-51.7$ & $[-65.0,-38.3]$ & $1.742\times10^{-7}$\\

Qwen / 4096 & F $\to$ A & 44 & 42 & 0/2 & $-3.3$ & $[-8.3,0.0]$ & $1$\\

Qwen / 4096 & F $\to$ W & 44 & 46 & 7/5 & $+3.3$ & $[-8.3,15.0]$ & $1$\\

Qwen / 4096 & C $\to$ A & 13 & 42 & 31/2 & $+48.3$ & $[33.3,61.7]$ & $2.486\times10^{-6}$\\

Qwen / 4096 & C $\to$ W & 13 & 46 & 34/1 & $+55.0$ & $[41.7,66.7]$ & $4.820\times10^{-8}$\\

Qwen / 4096 & A $\to$ W & 42 & 46 & 9/5 & $+6.7$ & $[-5.0,18.3]$ & $1$\\

\bottomrule\end{tabular}

\captionof{table}{All 24 registered primary comparisons on 60 paired questions each. F: full; C: concise; A: full then answer; W: warned then answer. One Holm family spans all cells and pairs. Intervals are pointwise stratified bootstrap intervals, not simultaneous intervals. Suffix-only and whole-trace sensitivity families give identical matching results; all 72 comparisons are supplied in the artifact.}\label{tab:answer_phase_all}\end{minipage}\end{lrbox}

At 2,048 tokens, Phi moves from 38 full matches to 50/51 with A/W, with 13 rescues and 1/0 harms. Both improvements pass the registered correction. Concise gives 41 matches; its contrasts with A/W have Holm $p=.2920/.0952$, so neither establishes superiority over concise. The full-to-A/W contrasts do not pass correction at 4,096. Qwen rarely requires continuation, and its controls show no significant difference from full after correction; this leaves equivalence unresolved.

\twocolumn[{\noindent\usebox{\comparisonledger}\par\medskip
\noindent\begin{minipage}{\textwidth}\centering\small

\begin{tabular}{llrrrrrr}

\toprule

Deployment / cap & Arm & Calls & Two-phase & Output tok. & Input tok. & Time (s) & Median tok.\\\midrule

Phi / 2048 & F & 60 & 0 & 88991 & 6393 & 1908.4 & 1498.0\\

Phi / 2048 & C & 60 & 0 & 70200 & 6513 & 1487.6 & 967.0\\

Phi / 2048 & A & 86 & 26 & 88588 & 59556 & 1910.8 & 1575.5\\

Phi / 2048 & W & 89 & 29 & 86728 & 67369 & 1876.2 & 1783.5\\

Phi / 4096 & F & 60 & 0 & 127810 & 6393 & 2757.9 & 1549.5\\

Phi / 4096 & C & 60 & 0 & 101592 & 6513 & 2199.5 & 1073.5\\

Phi / 4096 & A & 74 & 14 & 127568 & 63739 & 2759.8 & 1645.0\\

Phi / 4096 & W & 78 & 18 & 133237 & 81463 & 2892.5 & 1892.0\\

Qwen / 2048 & F & 60 & 0 & 35175 & 6440 & 1023.3 & 405.0\\

Qwen / 2048 & C & 60 & 0 & 892 & 6560 & 39.7 & 4.0\\

Qwen / 2048 & A & 60 & 0 & 36170 & 6440 & 1061.0 & 411.5\\

Qwen / 2048 & W & 61 & 1 & 30941 & 9720 & 905.6 & 355.5\\

Qwen / 4096 & F & 60 & 0 & 36373 & 6440 & 1059.3 & 414.5\\

Qwen / 4096 & C & 60 & 0 & 1013 & 6560 & 42.9 & 4.0\\

Qwen / 4096 & A & 60 & 0 & 36197 & 6440 & 1052.7 & 405.0\\

Qwen / 4096 & W & 61 & 1 & 34560 & 11767 & 1003.2 & 347.0\\

\bottomrule\end{tabular}

\captionof{table}{Complete answer-phase cost ledger, 60 items per row. Each item remains within its generated-token cap across both stages. Injected prefixes consume input, not output, and extra calls/prefill mean this is not an equal-total-compute comparison. Input counts do not estimate uncached FLOPs.}\label{tab:answer_phase_cost}\par\medskip\normalsize\raggedright

The run completed 960 logical outcomes through 1,049 calls and 1,036,035 generated tokens. Frozen requests, runtime-source and scorer hashes, stage payloads, all three endpoints, exact tests, Holm correction, intervals and costs were checked. The complete artifact retains all attempts, calls and outcomes. These generations have not received the separate human review of the original 120 responses.
\end{minipage}}]

%% file: paper.bbl
\begin{thebibliography}{38}
\providecommand{\natexlab}[1]{#1}

\bibitem[{Aytes et~al.(2025)Aytes, Baek, and Hwang}]{aytes-etal-2025-sketch}
Simon~A. Aytes, Jinheon Baek, and Sung~Ju Hwang. 2025.
\newblock \href {https://doi.org/10.18653/v1/2025.emnlp-main.1236}
  {Sketch-of-thought: Efficient {LLM} reasoning with adaptive
  cognitive-inspired sketching}.
\newblock In \emph{Proceedings of the 2025 Conference on Empirical Methods in
  Natural Language Processing}, pages 24296--24320, Suzhou, China. Association
  for Computational Linguistics.

\bibitem[{Cobbe et~al.(2021)Cobbe, Kosaraju, Bavarian, Chen, Jun, Kaiser,
  Plappert, Tworek, Hilton, Nakano, Hesse, and
  Schulman}]{cobbe2021trainingverifierssolvemath}
Karl Cobbe, Vineet Kosaraju, Mohammad Bavarian, Mark Chen, Heewoo Jun, Lukasz
  Kaiser, Matthias Plappert, Jerry Tworek, Jacob Hilton, Reiichiro Nakano,
  Christopher Hesse, and John Schulman. 2021.
\newblock \href {https://arxiv.org/abs/2110.14168} {Training verifiers to solve
  math word problems}.
\newblock \emph{Preprint}, arXiv:2110.14168.

\bibitem[{Efron(1979)}]{efron1979}
B.~Efron. 1979.
\newblock \href {https://doi.org/10.1214/aos/1176344552} {Bootstrap methods:
  Another look at the jackknife}.
\newblock \emph{The Annals of Statistics}, 7.

\bibitem[{Fagerland et~al.(2013)Fagerland, Lydersen, and Laake}]{fagerland2013}
Morten~W Fagerland, Stian Lydersen, and Petter Laake. 2013.
\newblock \href {https://doi.org/10.1186/1471-2288-13-91} {The {McNemar} test
  for binary matched-pairs data: Mid-p and asymptotic are better than exact
  conditional}.
\newblock \emph{BMC Medical Research Methodology}, 13:91.

\bibitem[{Han et~al.(2025)Han, Wang, Fang, Zhao, Ma, and
  Chen}]{han-etal-2025-token}
Tingxu Han, Zhenting Wang, Chunrong Fang, Shiyu Zhao, Shiqing Ma, and Zhenyu
  Chen. 2025.
\newblock \href {https://doi.org/10.18653/v1/2025.findings-acl.1274}
  {Token-budget-aware {LLM} reasoning}.
\newblock In \emph{Findings of the Association for Computational Linguistics:
  ACL 2025}, pages 24842--24855, Vienna, Austria. Association for Computational
  Linguistics.

\bibitem[{Hendrycks et~al.(2021)Hendrycks, Burns, Kadavath, Arora, Basart,
  Tang, Song, and Steinhardt}]{hendrycks2021math}
Dan Hendrycks, Collin Burns, Saurav Kadavath, Akul Arora, Steven Basart, Eric
  Tang, Dawn Song, and Jacob Steinhardt. 2021.
\newblock \href
  {https://datasets-benchmarks-proceedings.neurips.cc/paper_files/paper/2021/file/be83ab3ecd0db773eb2dc1b0a17836a1-Paper-round2.pdf}
  {Measuring mathematical problem solving with the math dataset}.
\newblock In \emph{Proceedings of the Neural Information Processing Systems
  Track on Datasets and Benchmarks}, volume~1.

\bibitem[{Holm(1979)}]{holm1979}
S.~Holm. 1979.
\newblock \href {https://www.jstor.org/stable/4615733} {A simple sequentially
  rejective multiple test procedure}.
\newblock \emph{Scandinavian Journal of Statistics}, 6:65--70.

\bibitem[{Hua et~al.(2025)Hua, Tang, Gu, Gu, Wong, and
  Qin}]{hua-etal-2025-flaw}
Andong Hua, Kenan Tang, Chenhe Gu, Jindong Gu, Eric Wong, and Yao Qin. 2025.
\newblock \href {https://doi.org/10.18653/v1/2025.emnlp-main.1006} {Flaw or
  artifact? rethinking prompt sensitivity in evaluating {LLM}s}.
\newblock In \emph{Proceedings of the 2025 Conference on Empirical Methods in
  Natural Language Processing}, pages 19889--19899, Suzhou, China. Association
  for Computational Linguistics.

\bibitem[{Huang et~al.(2024)Huang, Chen, Mishra, Zheng, Yu, Song, and
  Zhou}]{ICLR2024_8b4add8b}
Jie Huang, Xinyun Chen, Swaroop Mishra, Huaixiu~Steven Zheng, Adams Yu, Xinying
  Song, and Denny Zhou. 2024.
\newblock \href
  {https://proceedings.iclr.cc/paper_files/paper/2024/file/8b4add8b0aa8749d80a34ca5d941c355-Paper-Conference.pdf}
  {Large language models cannot self-correct reasoning yet}.
\newblock In \emph{International Conference on Learning Representations},
  volume 2024, pages 32808--32824.

\bibitem[{Jin et~al.(2024)Jin, Yu, Shu, Zhao, Hua, Meng, Zhang, and
  Du}]{jin-etal-2024-impact}
Mingyu Jin, Qinkai Yu, Dong Shu, Haiyan Zhao, Wenyue Hua, Yanda Meng, Yongfeng
  Zhang, and Mengnan Du. 2024.
\newblock \href {https://doi.org/10.18653/v1/2024.findings-acl.108} {The impact
  of reasoning step length on large language models}.
\newblock In \emph{Findings of the Association for Computational Linguistics:
  ACL 2024}, pages 1830--1842, Bangkok, Thailand. Association for Computational
  Linguistics.

\bibitem[{Lee et~al.(2025)Lee, Che, and Peng}]{lee2025tokencomplexity}
Ayeong Lee, Ethan Che, and Tianyi Peng. 2025.
\newblock \href {https://doi.org/10.48550/arXiv.2503.01141} {How well do {LLMs}
  compress their own chain-of-thought? a token complexity approach}.
\newblock \emph{Preprint}, arXiv:2503.01141.

\bibitem[{Lee et~al.(2026)Lee, D'Antoni, and
  Berg-Kirkpatrick}]{lee2026formattax}
Ivan~Yee Lee, Loris D'Antoni, and Taylor Berg-Kirkpatrick. 2026.
\newblock \href {https://arxiv.org/abs/2604.03616} {The format tax}.
\newblock \emph{Preprint}, arXiv:2604.03616.

\bibitem[{Li et~al.(2026)Li, Zhao, Zhang, and Gan}]{li-etal-2026-steering}
Junyan Li, Wenshuo Zhao, Yang Zhang, and Chuang Gan. 2026.
\newblock \href {https://doi.org/10.18653/v1/2026.findings-acl.1866} {Steering
  {LLM} thinking with budget guidance}.
\newblock In \emph{Findings of the {A}ssociation for {C}omputational
  {L}inguistics: {ACL} 2026}, pages 37437--37451, San Diego, California, United
  States. Association for Computational Linguistics.

\bibitem[{Lightman et~al.(2024)Lightman, Kosaraju, Burda, Edwards, Baker, Lee,
  Leike, Schulman, Sutskever, and Cobbe}]{ICLR2024_aca97732}
Hunter Lightman, Vineet Kosaraju, Yuri Burda, Harrison Edwards, Bowen Baker,
  Teddy Lee, Jan Leike, John Schulman, Ilya Sutskever, and Karl Cobbe. 2024.
\newblock \href
  {https://proceedings.iclr.cc/paper_files/paper/2024/file/aca97732e30bcf1303bc22ac3924fd16-Paper-Conference.pdf}
  {Let\textquotesingle s verify step by step}.
\newblock In \emph{International Conference on Learning Representations},
  volume 2024, pages 39578--39601.

\bibitem[{Lu et~al.(2024)Lu, Schuff, and Gurevych}]{lu-etal-2024-prompts}
Sheng Lu, Hendrik Schuff, and Iryna Gurevych. 2024.
\newblock \href {https://doi.org/10.18653/v1/2024.naacl-long.325} {How are
  prompts different in terms of sensitivity?}
\newblock In \emph{Proceedings of the 2024 Conference of the North American
  Chapter of the Association for Computational Linguistics: Human Language
  Technologies (Volume 1: Long Papers)}, pages 5833--5856, Mexico City, Mexico.
  Association for Computational Linguistics.

\bibitem[{Madaan et~al.(2023)Madaan, Tandon, Gupta, Hallinan, Gao, Wiegreffe,
  Alon, Dziri, Prabhumoye, Yang, Gupta, Majumder, Hermann, Welleck,
  Yazdanbakhsh, and Clark}]{NEURIPS2023_91edff07}
Aman Madaan, Niket Tandon, Prakhar Gupta, Skyler Hallinan, Luyu Gao, Sarah
  Wiegreffe, Uri Alon, Nouha Dziri, Shrimai Prabhumoye, Yiming Yang, Shashank
  Gupta, Bodhisattwa~Prasad Majumder, Katherine Hermann, Sean Welleck, Amir
  Yazdanbakhsh, and Peter Clark. 2023.
\newblock \href {https://doi.org/10.52202/075280-2019} {Self-refine: Iterative
  refinement with self-feedback}.
\newblock In \emph{Advances in Neural Information Processing Systems},
  volume~36, pages 46534--46594. Curran Associates, Inc.

\bibitem[{McNemar(1947)}]{mcnemar1947}
Quinn McNemar. 1947.
\newblock \href {https://doi.org/10.1007/BF02295996} {Note on the sampling
  error of the difference between correlated proportions or percentages}.
\newblock \emph{Psychometrika}, 12:153--157.

\bibitem[{Mizrahi et~al.(2024)Mizrahi, Kaplan, Malkin, Dror, Shahaf, and
  Stanovsky}]{mizrahi-etal-2024-state}
Moran Mizrahi, Guy Kaplan, Dan Malkin, Rotem Dror, Dafna Shahaf, and Gabriel
  Stanovsky. 2024.
\newblock \href {https://doi.org/10.1162/tacl_a_00681} {State of what art? a
  call for multi-prompt {LLM} evaluation}.
\newblock \emph{Transactions of the Association for Computational Linguistics},
  12:933--949.

\bibitem[{Molfese et~al.(2025)Molfese, Moroni, Gioffr{\'e}, Scir{\`e}, Conia,
  and Navigli}]{molfese-etal-2025-right}
Francesco~Maria Molfese, Luca Moroni, Luca Gioffr{\'e}, Alessandro Scir{\`e},
  Simone Conia, and Roberto Navigli. 2025.
\newblock \href {https://doi.org/10.18653/v1/2025.findings-acl.950} {Right
  answer, wrong score: Uncovering the inconsistencies of {LLM} evaluation in
  multiple-choice question answering}.
\newblock In \emph{Findings of the Association for Computational Linguistics:
  ACL 2025}, pages 18477--18494, Vienna, Austria. Association for Computational
  Linguistics.

\bibitem[{Panda et~al.(2026)Panda, Kadasi, Upperwal, and
  Singh}]{panda2026procedural}
Sailesh Panda, Pritam Kadasi, Abhishek Upperwal, and Mayank Singh. 2026.
\newblock \href {https://doi.org/10.48550/arXiv.2605.00817} {When {LLMs} stop
  following steps: A diagnostic study of arithmetic procedural execution in
  language models}.
\newblock \emph{Preprint}, arXiv:2605.00817.

\bibitem[{Pandey et~al.(2026)Pandey, Ye, and
  Li}]{pandey2026answerdecodingbehaviorllms}
Rohan Pandey, Eric Ye, and Michael Li. 2026.
\newblock \href {https://arxiv.org/abs/2603.28038} {Beyond the answer: Decoding
  the behavior of llms as scientific reasoners}.
\newblock \emph{Preprint}, arXiv:2603.28038.

\bibitem[{Renze and Guven(2024)}]{renze2024concise}
Matthew Renze and Erhan Guven. 2024.
\newblock \href {https://doi.org/10.1109/FLLM63129.2024.10852493} {The benefits
  of a concise chain of thought on problem-solving in large language models}.
\newblock In \emph{2nd International Conference on Foundation and Large
  Language Models (FLLM)}, pages 476--483.

\bibitem[{Schlesinger et~al.(2026)Schlesinger, Kim, Di~Martino, and
  Sapiro}]{schlesinger-etal-2026-claro}
Oded Schlesinger, Young~Kyung Kim, J.~Matias Di~Martino, and Guillermo Sapiro.
  2026.
\newblock \href {https://doi.org/10.18653/v1/2026.findings-acl.1335} {{CLARO}:
  Controlled attribute-driven reasoning optimization for efficient
  chain-of-thought}.
\newblock In \emph{Findings of the {A}ssociation for {C}omputational
  {L}inguistics: {ACL} 2026}, pages 26779--26799, San Diego, California, United
  States. Association for Computational Linguistics.

\bibitem[{Sclar et~al.(2024)Sclar, Choi, Tsvetkov, and
  Suhr}]{ICLR2024_6c0e99d7}
Melanie Sclar, Yejin Choi, Yulia Tsvetkov, and Alane Suhr. 2024.
\newblock \href
  {https://proceedings.iclr.cc/paper_files/paper/2024/file/6c0e99d736da621403018ca7b32b1a4d-Paper-Conference.pdf}
  {Quantifying language models\textquotesingle sensitivity to spurious features
  in prompt design or: How i learned to start worrying about prompt
  formatting}.
\newblock In \emph{International Conference on Learning Representations},
  volume 2024, pages 25055--25083.

\bibitem[{Sun et~al.(2026)Sun, Cheng, Li, Chen, and Wang}]{sun-etal-2026-stop}
Renliang Sun, Wei Cheng, Dawei Li, Haifeng Chen, and Wei Wang. 2026.
\newblock \href {https://doi.org/10.18653/v1/2026.acl-long.1256} {Stop when
  enough: Adaptive early-stopping for chain-of-thought reasoning}.
\newblock In \emph{Proceedings of the 64th Annual Meeting of the {A}ssociation
  for {C}omputational {L}inguistics (Volume 1: Long Papers)}, pages
  27250--27268, San Diego, California, United States. Association for
  Computational Linguistics.

\bibitem[{Sun et~al.(2025)Sun, Wang, Li, Liu, Li, Wen, Yuan, Zheng, Liang, Li,
  and Liu}]{sun-etal-2025-empirical}
Yi~Sun, Han Wang, Jiaqiang Li, Jiacheng Liu, Xiangyu Li, Hao Wen, Yizhen Yuan,
  Huiwen Zheng, Yan Liang, Yuanchun Li, and Yunxin Liu. 2025.
\newblock \href {https://doi.org/10.18653/v1/2025.emnlp-main.389} {An empirical
  study of {LLM} reasoning ability under strict output length constraint}.
\newblock In \emph{Proceedings of the 2025 Conference on Empirical Methods in
  Natural Language Processing}, pages 7652--7671, Suzhou, China. Association
  for Computational Linguistics.

\bibitem[{Turpin et~al.(2023)Turpin, Michael, Perez, and
  Bowman}]{NEURIPS2023_ed3fea90}
Miles Turpin, Julian Michael, Ethan Perez, and Samuel Bowman. 2023.
\newblock \href {https://doi.org/10.52202/075280-3275} {Language models
  don\textquotesingle t always say what they think: Unfaithful explanations in
  chain-of-thought prompting}.
\newblock In \emph{Advances in Neural Information Processing Systems},
  volume~36, pages 74952--74965. Curran Associates, Inc.

\bibitem[{Wang et~al.(2026)Wang, Zhang, Cao, Dou, Luo, MA, Zhu, Li, Huang, and
  Li}]{wang-etal-2026-format}
Dingzirui Wang, Xuanliang Zhang, Rongyu Cao, Longxu Dou, Xianzhen Luo, Yingwei
  MA, Qingfu Zhu, Binhua Li, Fei Huang, and Yongbin Li. 2026.
\newblock \href {https://doi.org/10.18653/v1/2026.findings-acl.1124}
  {Format-adapter: Improving reasoning capability of {LLM}s by adapting
  suitable format}.
\newblock In \emph{Findings of the {A}ssociation for {C}omputational
  {L}inguistics: {ACL} 2026}, pages 22408--22427, San Diego, California, United
  States. Association for Computational Linguistics.

\bibitem[{Wang et~al.(2024)Wang, Jain, Zhang, Ray, Kumar, and
  Athiwaratkun}]{wang-etal-2024-reasoning-token}
Junlin Wang, Siddhartha Jain, Dejiao Zhang, Baishakhi Ray, Varun Kumar, and Ben
  Athiwaratkun. 2024.
\newblock \href {https://doi.org/10.18653/v1/2024.emnlp-main.1112} {Reasoning
  in token economies: Budget-aware evaluation of {LLM} reasoning strategies}.
\newblock In \emph{Proceedings of the 2024 Conference on Empirical Methods in
  Natural Language Processing}, pages 19916--19939, Miami, Florida, USA.
  Association for Computational Linguistics.

\bibitem[{Wang et~al.(2023)Wang, Wei, Schuurmans, Le, Chi, Narang, Chowdhery,
  and Zhou}]{wang2023selfconsistency}
Xuezhi Wang, Jason Wei, Dale Schuurmans, Quoc Le, Ed~H. Chi, Sharan Narang,
  Aakanksha Chowdhery, and Denny Zhou. 2023.
\newblock \href {https://arxiv.org/abs/2203.11171} {Self-consistency improves
  chain of thought reasoning in language models}.
\newblock In \emph{International Conference on Learning Representations}.

\bibitem[{Wang and Zhou(2024)}]{NEURIPS2024_7a8e7fd2}
Xuezhi Wang and Denny Zhou. 2024.
\newblock \href {https://doi.org/10.52202/079017-2123} {Chain-of-thought
  reasoning without prompting}.
\newblock In \emph{Advances in Neural Information Processing Systems},
  volume~37, pages 66383--66409. Curran Associates, Inc.

\bibitem[{Wang et~al.(2025)Wang, Liu, Wang, Li, Wei, Liu, and
  Bao}]{wang2025promptbridgecrossmodelprompttransfer}
Yaxuan Wang, Quan Liu, Zhenting Wang, Zichao Li, Wei Wei, Yang Liu, and Yujia
  Bao. 2025.
\newblock \href {https://arxiv.org/abs/2512.01420} {Promptbridge: Cross-model
  prompt transfer for large language models}.
\newblock \emph{Preprint}, arXiv:2512.01420.

\bibitem[{Wei et~al.(2022)Wei, Wang, Schuurmans, Bosma, Ichter, Xia, Chi, Le,
  and Zhou}]{NEURIPS2022_9d560961}
Jason Wei, Xuezhi Wang, Dale Schuurmans, Maarten Bosma, Brian Ichter, Fei Xia,
  Ed~Chi, Quoc~V Le, and Denny Zhou. 2022.
\newblock \href {https://doi.org/10.52202/068431-1800} {Chain-of-thought
  prompting elicits reasoning in large language models}.
\newblock In \emph{Advances in Neural Information Processing Systems},
  volume~35, pages 24824--24837. Curran Associates, Inc.

\bibitem[{Xu et~al.(2025)Xu, Xie, Zhao, and
  He}]{xu2025chaindraftthinkingfaster}
Silei Xu, Wenhao Xie, Lingxiao Zhao, and Pengcheng He. 2025.
\newblock \href {https://arxiv.org/abs/2502.18600} {Chain of draft: Thinking
  faster by writing less}.
\newblock \emph{Preprint}, arXiv:2502.18600.

\bibitem[{Yang et~al.(2026)Yang, Wang, Ganguly, Li, Song, Singh, Chaudhary, and
  Han}]{yang-etal-2026-mid}
Van Yang, Shouren Wang, Debargha Ganguly, Xinpeng Li, Chaoda Song, Vikash
  Singh, Vipin Chaudhary, and Xiaotian Han. 2026.
\newblock \href {https://doi.org/10.18653/v1/2026.findings-acl.299} {Mid-think:
  Training-free intermediate-budget reasoning via token-level triggers}.
\newblock In \emph{Findings of the {A}ssociation for {C}omputational
  {L}inguistics: {ACL} 2026}, pages 6024--6038, San Diego, California, United
  States. Association for Computational Linguistics.

\bibitem[{Yuan et~al.(2024)Yuan, Shen, Yan, Zhang, Xie, Wang, Guan, Wang, and
  Ye}]{NEURIPS2024_e304e04a}
Xiaosong Yuan, Chen Shen, Shaotian Yan, Xiaofeng Zhang, Liang Xie, Wenxiao
  Wang, Renchu Guan, Ying Wang, and Jieping Ye. 2024.
\newblock \href {https://doi.org/10.52202/079017-3986} {Instance-adaptive
  zero-shot chain-of-thought prompting}.
\newblock In \emph{Advances in Neural Information Processing Systems},
  volume~37, pages 125469--125486. Curran Associates, Inc.

\bibitem[{Zhou et~al.(2026)Zhou, Ling, Chen, Wang, Fan, and
  Wang}]{zhou-etal-2026-thinking}
Shu Zhou, Rui Ling, Junan Chen, Xin Wang, Tao Fan, and Hao Wang. 2026.
\newblock \href {https://doi.org/10.18653/v1/2026.findings-acl.1199} {When more
  thinking hurts: Overthinking in {LLM} test-time compute scaling}.
\newblock In \emph{Findings of the {A}ssociation for {C}omputational
  {L}inguistics: {ACL} 2026}, pages 23967--23977, San Diego, California, United
  States. Association for Computational Linguistics.

\bibitem[{Zhuo et~al.(2024)Zhuo, Zhang, Fang, Duan, Lin, and
  Chen}]{zhuo-etal-2024-prosa}
Jingming Zhuo, Songyang Zhang, Xinyu Fang, Haodong Duan, Dahua Lin, and Kai
  Chen. 2024.
\newblock \href {https://doi.org/10.18653/v1/2024.findings-emnlp.108}
  {{P}ro{SA}: Assessing and understanding the prompt sensitivity of {LLM}s}.
\newblock In \emph{Findings of the Association for Computational Linguistics:
  EMNLP 2024}, pages 1950--1976, Miami, Florida, USA. Association for
  Computational Linguistics.

\end{thebibliography}
